\documentclass[journal]{IEEEtran}

\usepackage{lastpage} % Required to determine the last page for the footer
\usepackage{amssymb}
\usepackage{color}
\usepackage{amsmath}
\usepackage{amsthm}
\usepackage{amsmath}
\usepackage{amsfonts}
\usepackage{url}
\usepackage{multirow}
\usepackage{footnote}
\usepackage{amssymb}
\usepackage{mathtools}
\usepackage{algorithmicx}
\usepackage{algorithm}
\usepackage{algpseudocode} 
\usepackage{cases}
\usepackage{cite}
\usepackage{arydshln}
\usepackage{makecell}
\usepackage{tabularx}
\usepackage[table]{colortbl}
\usepackage{booktabs}
\usepackage{tikz}
\usetikzlibrary{arrows,shapes,fit}
\usepackage{caption}
\usepackage{subcaption}
\usepackage{subfig}
\usepackage{dblfloatfix} 
\usepackage{flushend}
\usepackage{standalone}
\usepackage{soul,xcolor}
\usepackage{color, colortbl}
\usepackage{rotating}
\usepackage{blindtext}
\usepackage{pgfplots}
\pgfplotsset{compat=newest}
\usepackage[thinlines]{easytable}
\usepackage{enumitem}
\usepackage{ragged2e}
\usepackage{titlesec}
\titlespacing*{\subsection}{0pt}{.1\baselineskip}{.1\baselineskip}
\titlespacing*{\section}{0pt}{.2\baselineskip}{.2\baselineskip}
\input{mystyle/mysymbol.sty}
\usepackage{needspace}

\input{mystyle/mycolors.sty}

\begin{document}

\title{GKNet: Graph Kalman Filtering and Model Inference via Model-based Deep Learning}
\author{Mohammad Sabbaqi, Riccardo Taormina and Elvin Isufi\vspace{-.75cm}
\IEEEcompsocitemizethanks{\IEEEcompsocthanksitem
The authors are with the Delft University of Technology, Delft, The Netherlands; 
E-mails: \{m.sabbaqi, r.taormina, e.isufi-1\}@tudelft.nl.
Part of this work has been presented in~\cite{sabbaqi_inferring}.
This work is in part supported by the TU Delft AI Labs program and by the TTW-OTP project GraSPA (project number 19497) financed by the Dutch Research Council (NWO).
}%
\thanks{}}

\maketitle

\begin{abstract}
\justifying{
Inference tasks with time series over graphs are of importance in applications such as urban water networks, economics, and networked neuroscience. Addressing these tasks typically relies on identifying a computationally affordable model that jointly captures the graph-temporal patterns of the data. In this work, we propose a graph-aware state space model for graph time series, where both the latent state and the observation equation are parametric graph-induced models with a limited number of parameters that need to be learned. More specifically, we consider the state equation to follow a stochastic partial differential equation driven by noise over the graphs’ edges accounting not only for potential edge uncertainties but also for increasing the degrees of freedom in the latter in a tractable manner. The graph structure conditioning of the noise dispersion allows the state variable to deviate from the stochastic process in certain neighborhoods. The observation model is a sampled and graph-filtered version of the state capturing multi-hop neighboring influence. The goal is to learn the parameters in both state and observation models from the partially observed data for downstream tasks such as prediction and imputation. The model is inferred first through a maximum likelihood approach that provides theoretical tractability but is limited in expressivity and scalability. To improve on the latter, we use the state-space formulation to build a principled deep learning architecture that jointly learns the parameters and tracks the state in an end-to-end manner in the spirit of Kalman neural networks. The proposed models are evaluated on controlled and four real-world scenarios, showing the effectiveness of the proposed approach when the model lacks training data. Finally, we conduct a case study on water networks, where the proposed model is used to predict the water levels in the network when the data is extremely scarce.
}
\end{abstract}

%%%---------------------------------------------------------------------
\section{Introduction}
\label{sec_intro}
% ---------------------------------------------------------------------------------------
% INTRODUCTION
% ---------------------------------------------------------------------------------------

% ---------------------------------------------------------------------------------------
% motivation
\IEEEPARstart{I}{nference} of time-varying signals over graphs plays a key role in network-based systems to interpolate missing values, extrapolate a certain horizon, and detect anomalies, to name a few~\cite{kolaczyk_statistical_2014,ortega_overview_2018,marisca_learning_2022,yi_fouriergnn_nodate,li_deep_2023}.
This is a challenging problem because the relatio-temporal coupling in the data needs to be exploited in a computationally affordable manner due to the problem dimensions involving large graphs and long time horizons.
The challenge increases when the graph is imperfect, which is often the case in almost all physical networks (water, power, transport) but also when the said graph is estimated from a finite amount of data~\cite{giannakis_topology_2018,dong_learning_2019,mateos_connecting_2019,buciulea_learning_2022,shen_kernel_2017}.
The impact of the topological uncertainties has been studied in different settings including graph filtering~\cite{isufi_graphfilters_2024}, graph signal processing~\cite{ceci_graph_2020}, graph neural networks \cite{sabbaqi_gtcnn,gao_learning_2023,gao_stochastic_2021,gao_stability_2021}, and PDE-based graph machine learning~\cite{chamberlain_grand_2021,song_robustness_2022}.

% ---------------------------------------------------------------------------------------
% Related work: statistical
A natural way to infer time-varying signals on graphs is to extend graph kernels~\cite{smola_kernels_2003} into a relatio-temporal form, where a separable kernel is favored due to computational aspects~\cite{romero_kernelbased_2017,lu_probabilistic_2021}.
Another approach is to design a latent space model where independent Gaussian processes pass through a graph filter to account for both the temporal and the spatial connections in the data~\cite{lu_spatiotemporal_2021,zhi_gaussian_2023}.
The work in~\cite{nikitin_non_2022} also builds around Gaussian processes but designs a joint (non-separable) kernel based on stochastic partial differential equations~(SPDEs).
The latter is more interpretable and expressive than the separable kernels, but it is computationally heavier.
In general, these statistical approaches are limited in scalability, and rely on statistical assumptions that may not hold.

% ---------------------------------------------------------------------------------------
% Related work: machine learning
Another approach to model time-varying signals over graphs is to use temporal variants of graph neural networks (GNNs).
One stream of works focuses on sequentially combining GNNs with recurrent neural networks or their variants~\cite{jin_survey_2024}.
% The works in~\cite{chai_bike_2018,manessi_dynamic_2020,sun_constructing_2021,khodayar_spatio-temporal_2019} concatenate a simple GNN with an LSTM network where the former functions node-wise and the latter captures the temporal dependencies.
% Graph WaveNet uses leverages gated dilated temporal convolutions to better capture long-term patterns and a layer of graph convolution accounts for the spatial dependencies~\cite{wu_graph_2019}.
Another stream of works, related to this paper, designs graph neural networks that can capture both relational and temporal dependencies simultaneously.
For example, the authors in~\cite{ruiz_gated_2020} insert graph convolutions within a GRU architecture to consider spatial connections while exploiting temporal patterns.
% The work in~\cite{yan_spatial_2018} accounts for temporal immediate neighbors in the message-passing process to retrieve temporal information.
The work in~\cite{sabbaqi_gtcnn} propose an architecture performing on a product graph of spatial connections and temporal dependencies.
The main drawback of these approaches is that they require a large amount of data to learn the model parameters, and treat the graph structure as solely the domain where graph convolutions are performed, ultimately, limiting the interpretability of the process dynamics.
% Moreover, they do not consider the uncertainties in the graph structure in their deterministic predictions.

% ---------------------------------------------------------------------------------------
% Related work: model-based deep learning
To overcome the limitations of the above approaches, one needs to combine the interpretability and reliability of statistical approaches with the flexibility and scalability of machine learning approaches.
Model-based deep learning~\cite{shlezinger_model-based_2022,shlezinger_model-based_2023} is a promising direction in this regard, which aims to develop data-driven solutions where particular models or respective optimization problems are used as inductive biases to design principled architectures that allow a degree of tractability~\cite{hadou_robust_2023}.
Model-based deep learning for graphs has been proposed in~\cite{sabbaqi_gtfunrolling,chen_graph_2021,nagahama_graph_2022,chen_time_2021,vu_Unrolling_2021}.
However, these models are restricted to a certain task, typically time-invariant, and data.
The latent perspective is often key to model the time series evolution as the driving factors may not be observed~\cite{isufi_observing_2020}.
Moreover, they work either on static graph signals or short temporal windows.

State space model-based deep learning solutions are a powerful alternative in this regard.
In this context, the authors in~\cite{revach_kalman_2022} proposed KalmanNet for tracking state variable which is a deep learning architecture that learns the Kalman gain parameters via a recurrent neural network.
Later on,~\cite{sagi_extended_2023} proposed a spectral-based approach for extended Kalman filtering in non-linear graph state space models.
Based on this work, the GSP-KalmanNet~\cite{buchnik_gspkalman_2023} is proposed to track graph time-varying signals in state space models similar to KalmanNet.
However, they consider both the observation and transition models known which is rarely the case.

% ---------------------------------------------------------------------------------------
% Work positioning
In this work we propose a graph-based state space model for inferring time varying signals over graphs where both the state and observation models are unknown.
This is a high-dimensional ill-posed problem, and, to make it tractable, we consider the state equation based on a graph SPDE kernel that accounts for the uncertainties in the graph structure.
The observation model is a locally graph-filtered version of the latent graph signals to prevent over-smooth predictions. 
We use said parametric state space formulation to model the process evolution on two fronts.
First, both the state and observation models are learned through an optimization problem in a maximum likelihood setting alongside the latent time series.
This method is theoretically grounded and interpretable, but it is computationally heavy and lacks expressivity.
Thus, we extend the proposed algorithm into a model-based deep learning architecture with a recurrent inference model to enhance its robustness, expressive power, and scalability even in the presence of topological uncertainties and few data samples.
% ---------------------------------------------------------------------------------------
% contribution
Our main contributions are fourth fold:
\begin{itemize}
    \item \textbf{Graph-based state space model with uncertainties:} 
    We propose a graph-based state space model for inferring time-varying signals over graphs that accounts for uncertainties.
    This state model is inspired by topological SPDE-based kernels over graphs where the dispersion of the system noise is closely tied to the graph structure.
    Then, the state passes through a graph filter to obtain the observations.
    \item \textbf{Learning via Kalman filter and maximum likelihood:}
    We propose an inference algorithm for time series over graphs.
    The Kalman filter is used alternatively with a maximum likelihood estimation algorithm to learn the observation and state model parameters, as well as to infer the latent time series simultaneously, within an expectation maximization setting.
    \item \textbf{Model-based deep learning architecture:}
    We extend the proposed model into a model-based deep learning architecture to enhance its robustness, expressive power, and scalability.
    To do so, we use a recurrent recognition model to learn the model parameters and topological uncertainties in an end-to-end manner.
    \item \textbf{Numerical experiments:}
    Extensive experiments on tasks such as forecasting and time series interpolation over a synthetic and four real-world datasets demonstrate the effectiveness of the proposed model.
    The results show a favorable performance when compared with both statistical and machine learning approaches.
    % An ablation study is also conducted to show the importance of each component of the proposed model in different settings.
    \item \textbf{Water network systems case study:}
    A case study on water network systems is conducted to show the effectiveness of the proposed model in real-world applications beyond the benchmark datasets.
    The method is modified for the case study to account for the specific characteristics of the water network systems.
\end{itemize}

% ---------------------------------------------------------------------------------------
% paper organization

The rest of the paper is organized as follows.
In Section~\ref{sec_back}, we provide the background to formalize the problem.
Section~\ref{sec_em_inference} proposes an algorithm to infer the latent time series and learn the model parameters in a maximum likelihood setting based on Kalman filtering and EM algorithm.
In Section~\ref{sec_rgkn}, we extend the proposed model into a model-based deep learning architecture with a recurrent recognition model.
Section~\ref{sec_numeric} presents the experimental results on synthetic and real-world data.
Section~\ref{sec_casestudy} discusses a case study on water network systems.
Finally, we conclude the paper in Section~\ref{sec_conc}.

% ---------------------------------------------------------------------------------------
%  END: INTRODUCTION
% ---------------------------------------------------------------------------------------
%%%---------------------------------------------------------------------

%%%---------------------------------------------------------------------
\section{Problem Formulation}
\label{sec_back}
%-------------------------------------------------------------------
% background

% graph definitions
Consider an undirected graph $\ccalG = \{\ccalV,\ccalE,\ccalW\}$ with the node set $\ccalV$, the edge set $\ccalE$, and the edge weights $\ccalW$.
Let the graph $\ccalG$ have $|\ccalV| = N$ nodes and $|\ccalE| = M$ edges, and let it be represented by its adjacency matrix $\bbA \in \reals^{N\times N}$.
The Laplacian matrix $\bbL = \diag(\bbA\bbone) - \bbA$ is another representation matrix behaving similarly to a difference operator where $\bbone$ is an all-one vector and $\diag(\cdot)$ transforms a vector into a diagonal matrix.
Matrix $\bbB \in \reals^{N\times M}$ is the node-to-edge incidence matrix of graph $\ccalG$ which indicates the nodes connected by each edge.
One can show that the graph Laplacian matrix $\bbL = \bbB\bbB^\top$ while $\bbL_1 = \bbB^\top\bbB$ is defined as the edge Laplacian matrix.
A time series over the nodes of the graph $\ccalG$ is a sequence of data points $\bbx_t \in \reals^N$ at time $t$ with the $i$-th entry $x_{it}$ being time series evolving over the node $i$.

% times series over graphs
Our ultimate goal is to model the time series data $\bbx_t$ over graph $\ccalG$ for downstream tasks such as prediction, anomaly detection, and signal reconstruction.
Under the assumption that the graph $\ccalG$ plays a role in the time series generation process, or it captures the data structure, the temporal dependencies of $\bbx_t$ are related to it.
In the simplest case, the time series can be generated by a Markovian process over the graph where the data at time $t$ is a transformation of the data at time $t-1$ constrained to the graph structure as
\begin{equation}
	\bbx_t = f(\bbA,\bbx_{t-1}) + \bbw_t,
	\label{eq:markovian_process}
\end{equation}
where $f(\cdot)$ is an arbitrary function that depends on the adjacency matrix $\bbA$ and $\bbw_t$ is the noise term.
Identifying function $f(\cdot)$ from signal observations is the central problem in different disciplines as we elaborate next.

%-------------------------------------------------------------------
% problem motivation
\subsection{Problem Motivation}
\label{sec:problem_motivation}

% inferring evolutions over graphs
Modeling the time series over graphs is considered in social networks, transportation systems, and power grids where the data is collected over a network of nodes and the evolution is influenced by the network structure.
However, the observed data might not strictly follow the Markovian process in~\eqref{eq:markovian_process} and the evolution might be more complex.
Hence, modeling the data using such strong statistical assumptions might be insufficient to capture the underlying process dynamics.  

% graph state space models
State space models help to overcome this issue and consider the observed data $\bby_t$ is a local aggregation of a latent state variable $\bbx_t$ that evolves over the graph.
The state variable $\bbx_t$ follows a Markovian process influenced by the graph structure as in \eqref{eq:markovian_process}.
The observation model locally combines the state variable $\bbx_t$ considering the graph structure as
\begin{equation}
	\bby_t = g(\bbA,\bbx_t) + \bbv_t,
	\label{eq:observation_model}
\end{equation}
where $g(\cdot)$ is an arbitrary function and $\bbv_t$ is the observation noise.

% related works
Given the observation model $g(\cdot)$ and the state model $f(\cdot)$ are known, the state variable $\bbx_t$ can be estimated via graph Kalman filtering.
This problem was initially addressed in~\cite{shi_kalman_2009} for state estimation in sensor networks with distributed computations.
Authors in~\cite{isufi_observing_2020} proposed the graph Kalman filter for band-limited processes over graphs given the network structure and sampled measurements.
Later on, the unscented graph Kalman filter~\cite{li_unscented_2023} and the extended graph Kalman filter~\cite{sagi_extended_2023} were proposed to handle nonlinearities in the dynamics and observation model.
Moreover, the work in~\cite{su_graph-frequency_2024} proposed a graph-frequency domain Kalman filter to enhance the tracking performance subject to outlier measurements. 
However, all of these methods suffer from instability against model mismatches and require a good initialization of the state variable.
Authors in~\cite{buchnik_gspkalman_2023} proposed a model-based neural network called GSP-KalmanNet to tackle this issue by learning the Kalman gain via a recurrent neural network from the data itself.
This method is more robust to model mismatches and leverages the graph structure, to improve the estimation performance.
Yet, both the observation and state models $g(\cdot)$ and $f(\cdot)$ are assumed known.

% research gap and problem statement
\smallskip
\noindent\textbf{Observation model.}
Unlike the abovementioned works, we aim to learn both the state model $f(\cdot)$ and the observation model $g(\cdot)$ to map the time series data into a well-behaved latent space.
More specifically, we consider the observation model to be a linear local aggregation of the state variable $\bbx_t$ over the graph combined with the measurement noise as
\begin{equation}
	\bby_t 
	=\bbM\left(\sum_{k=0}^{K}h_k\bbL^k\bbx_t\right) + \bbv_t
	=\bbM\tilde{\bbH}(\bbL)\bbx_t + \bbv_t,
	\label{eq:observation_model_linear}
\end{equation}
where $\tilde{\bbH}(\bbL) = \sum_{k=0}^{K}h_k\bbL^k$ is a graph convolutional filter of order $K$ with coefficients $\bbh = [h_0,\dots,h_K]^\top$~\cite{isufi_graphfilters_2024}.
The masking matrix $\bbM \in \{0,1\}^{N_o \times N}$ accounts for missing values in the observation with $N_o$ as the number of observed nodes.
In the rest of the paper, we use $\bbH = \bbM\tilde{\bbH}(\bbL)$ as the observation model for notational convenience.

% state equation
\smallskip
\noindent\textbf{State model.}
We consider the linear approximation of heat diffusion over the graph as the evolution process in the latent space as
\begin{equation}
	d\bbx_t = -c\bbL\bbx_{t}dt
	\label{eq:heat_diffusion}
\end{equation}
where $c$ is the diffusivity factor.
This is motivated by the fact that the heat diffusion process is the simplest process over a graph and can be represented with a minimum number of parameters.
However, directly using~\eqref{eq:heat_diffusion} can be too strict as it limits the feasible space for the learnable filter parameters $\bbh$ in the observation model.
Moreover, the evolution process relies on the exact graph structure which might not be the case in real-world applications.
To account for the uncertainties in the graph structure, and to allow for a more flexible model, we propose a state space model where the state variable $\bbx_t$ follows a SPDE-based graph state equation.
One can define the stochastic heat diffusion process over the graph as~\cite{nikitin_non_2022}
\begin{equation}
	d\bbx_t = -c\bbL\bbx_t dt + \bbS d\bbbeta_t
	\label{eq:spde_on_graph}
\end{equation}
where $\bbbeta_t$ is an $F$-dimensional standard Brownian motion and $\bbS \in \reals^{N\times F}$ is the dispersion matrix that distributes the noise over the graph throughout the process.
The solution of this SPDE hovers around the deterministic heat diffusion process over graphs defined in~\eqref{eq:heat_diffusion} up to the covariance of the noise term $\bbbeta_t$.
The intensity of the noise term around each node is controlled by the dispersion matrix $\bbS$ through graph and time which allows the model to be arbitrarily close to a heat diffusion process at each node independently.
This property is crucial for modeling time series over graphs as it allows the model to be more flexible and generic.

However, using a general diffusion matrix $\bbS$ comes with drawbacks:
\begin{enumerate}
	\item The high degree of freedom leads to model overfitting.
	\item The interpretability of the model is lost as the system noise disperses without any structure.
	\item It is computationally expensive and memory demanding as the dimension of $\bbS$ is dependent on the problem size.
	\item The model is non-scalable and non-transferable to other graphs as $\bbS$ is tailored to a fixed structure. 
\end{enumerate}
Thus, we propose a local uncertainty model for the latent variable which considers the graph structure as an inductive bias in the noise dispersion matrix as
\begin{equation}
	\derv\bbx_t = -c\bbL\bbx_t\derv t + \bbB\diag(\bbalpha)\derv\bbbeta_t,
	\label{eq:state_model}
\end{equation}
where $\bbalpha \in \reals^M$ is the local uncertainty vector that allows the process $\bbx_t$ to deviate from a diffusion process around a certain edge $m$ if the corresponding $\alpha_m$ is high.
Discretizing the continuous-time state equation~\eqref{eq:state_model} via the first-order Euler method, we get the linear state model
\begin{equation}
	\bbx_{t+1} = -c\bbL \bbx_{t} + \bbB\diag(\bbalpha)\bbw_t,
	\label{eq:state_disc}
\end{equation}
where $\bbw_t \sim \ccalN(\bbzero,\bbI)$ is the standard additive noise term.
As~\eqref{eq:state_disc} suggests, a white Gaussian noise $\bbw_t$ is considered over the edges of the graph where the variance of each edge is controlled by the corresponding learnable parameter $\bbalpha$.
Hence, the parameter $\bbalpha$ can be considered as the edge uncertainty vector which allows the process $\bbx_t$ to deviate from a heat diffusion process around a certain edge proportional to the intensity of uncertainty $\bbalpha$ in that edge.
This property has two main advantages: first, it provides more
degrees of freedom for estimating the observation model $\bbH$ in~\eqref{eq:observation_model_linear} as the span of valid latent diffusions $\bbx_t$ is increased;
second, it allows the model to be more robust to the structural uncertainties in the graph as the corresponding uncertainty term $\bbalpha$ is learned from the data itself.
%-------------------------------------------------------------------

%-------------------------------------------------------------------
% remark: link with gp
The following remarks are in order to link the proposed state space model with Gaussian processes and to interpret the uncertainty modeling capabilities of the model.

\begin{remark}
Under initial Gaussian conditions, i.e., $\bbx_0 \sim \ccalN(\bbmu_0,\bbSigma_0)$, the solution of~\eqref{eq:spde_on_graph} is a Gaussian process of the form $\bbx_t \sim \ccalG\ccalP(\bbmu_t,\bbSigma_{t,s}(\bbL))$ \cite{nikitin_non_2022}.
The mean process $\bbmu_t$ is a heat diffusion on graphs $\bbmu_t = \exp(-c\bbL t) \bbmu_0$.
The covariance function is $\bbSigma_{t,s}(\bbL) = \bbV \bbC_{t,s} \bbV^\top$, where matrix $\bbC_{t,s}$ has as $(i,j)$-th entry
\begin{equation}
	[\bbC_{t,s}]_{ij}= \frac{[\bbV^\top\bbS\bbS^\top\bbV]_{ij}}{c(\lambda_i+\lambda_j)}
	(e^{-c\lambda_i|t-s|}-e^{-c(\lambda_i t +\lambda_j s)}).
	\label{eq:non_sep_kernel}
\end{equation}
This covariance matrix defines a relatio-temporal kernel that suppresses higher graph frequency content and imposes low-pass properties.
If the matrix $\bbS\bbS^\top$ is analytic in $\bbL$, this kernel shrinks into a spectral Gaussian process over the graph~\cite{zhi_gaussian_2023}.

This parallel shows that the kernel in~\eqref{eq:non_sep_kernel} has in general two main limitations.
First, considering a time series of length $T$, its complexity is of order cubic $\ccalO(N^3T^3)$, limiting its applicability to small graphs and time horizons.
Second, the Brownian motion diffuses over the network regardless of the topology via a general diffusion matrix $\bbS$.
We tackle the first issue by implementing the Gaussian process recursively in a state space model as in~\eqref{eq:observation_model_linear}; and the second issue by matching the dispersion matrix $\bbS$ with the graph sparsity as in~\eqref{eq:state_disc}. 
\end{remark}
%-------------------------------------------------------------------

%-------------------------------------------------------------------
% remark: uncertainty perspective
\begin{remark}
To further shed light on the graph uncertainty modeling capabilities of~\eqref{eq:state_disc}, we  can consider the observed graph $\bbL$ matches the support of the true graph $\bbL^\star$ which drives the diffusion, but its edge weights differ slightly because of an estimation error.
This could be seen as a relative estimation error~\cite{gama_stability_2020,sabbaqi_gtcnn,levie_transferability_2019}, and it is common in many applications involving physical networks such as water, transportation, and power networks.
This error can be modeled as
\begin{equation}
	\bbL^\star = \bbL + \bbB\diag(\bbe)\bbB^\top,
\end{equation}
where $\bbe \in \reals^M$ is a standard noise modeling the uncertainties over each edge independently.
A heat diffusion model over the underlying graph $\bbL^\star$ can then be written as
\begin{equation}
	\textstyle
	\derv \bbx_t = -c\bbL^\star\bbx_t\derv t 
	= -c\bbL\bbx_t\derv t - c\bbB\diag(\bbe)\bbB^\top\bbx_t\derv t,
\end{equation}
where process $\bbx_t$ is the superposition of the diffusion over the observed graph $\bbL$ and of the relative error graph $\bbB\diag(\bbe)\bbB^\top$.
Note that $\bbL^\star$, $\bbL$, and $\bbB\diag(\bbe)\bbB^\top$ share the same sparsity pattern.

This SPDE is time-variant because of the term $\bbalpha_t = \bbB^\top\bbx_t$.
Our proposal could be seen as defining $\bbalpha = \bbalpha_t$ to get the linear time-invariant (LTI)-SPDE.
Although this approximation is not exact, except in the converging limit where the graph signals are smooth, it allows us to interpret the model as an uncertainty model over the edges that links this work with graph perturbation models.
The LTI-SPDE reads as
\begin{equation}
	\derv\bbx_t = -c\bbL\bbx_t\derv t + \bbB\diag(\bbalpha)\derv\bbbeta_t,
\end{equation}
with $\derv \bbbeta_t = \bbw\derv t$.
This means that the Brownian motion $\bbbeta_t$ has independent entries with different energies $\bbalpha$ that depend on the graph signal differences $\bbB^\top\bbx_t$.
Thus, the uncertainty will play a bigger role on edges with a bigger signal difference.
In turn, this shows how we can generalize the heat diffusion process over graphs to allow for uncertainties in the structure.
\end{remark}
%-------------------------------------------------------------------

%-------------------------------------------------------------------
% problem statement
\subsection{Problem Statement}
\label{sec:problem_statement}

Given the partially observed data $\bby_t$ over $T$ consecutive timestamps, and considering the state space model
\begin{equation}
	\left\{
	\begin{aligned}
		& \bbx_t = -c\bbL\bbx_{t-1} + \bbB\diag(\bbalpha)\bbw_t, \\
		& \bby_t = \bbH\bbx_t + \bbv_t \coloneq \bbM\tbH(\bbL)\bbx_t + \bbv_t,
	\end{aligned}
	\right.
	\label{eq:state_space_model}
\end{equation}
with initial condition $\bbx_0 \sim \ccalN(\bbzero,\sigma_0^2\bbI)$, system noise $\bbw_t \sim \ccalN(\bbzero,\bbI)$, and observation noise $\bbv_t \sim \ccalN(\bbzero,\sigma^2\bbI)$, our goal is to estimate the filter parameters $\bbh \in \reals^{K+1}$ in $\tbH(\bbL)$ and edge uncertainties $\bbalpha \in \reals^M$ to recover the missing entries in the observation $\bby_t$ and perform other down-stream tasks such as forecasting or anomaly detection.

The challenge here is that the latent variable $\bbx_t$ is not directly observed and a trade-off between the uncertainty in the latent space $\bbalpha$ and the degrees of freedom for the observation model $\bbh$ should be considered.
This problem can be linked to hidden Markov models~\cite{cappe_hmm_2005}, but the graph structure adds an extra layer of complexity.
We solve this problem in two ways.
First, by maximizing the likelihood function obtained from the solution of the state space model~\eqref{eq:state_space_model} through the expectation-maximization (EM) algorithm as it performs comparatively well when the number of observations in~\eqref{eq:state_space_model} is low~\cite{dempster_maximum_1977,baum_statistical_1966}.
However, in some cases, the statistical approaches are limited by the linearity of the model, strong statistical assumptions on the data, and scalability.
To overcome the latter, we build on the developed EM algorithm and propose a model-based deep learning approach to address these issues at the price of theoretical guarantees and interpretability.
%-------------------------------------------------------------------

%%%---------------------------------------------------------------------

%%%---------------------------------------------------------------------
\section{Learning Graph State-Space Model}
\label{sec_em_inference}
In this section, we derive an algorithm for learning the graph state-space model parameters based on the expectation-maximization framework.

%-------------------------------------------------------------------
% likelihood
% \vspace{-.5cm}
\subsection{Maximum Likelihood Estimation}
The joint likelihood of~\eqref{eq:state_space_model} is the joint probability of observations $\{\bby_1,\dots,\bby_T\}$ and states $\{\bbx_0,\dots,\bbx_T\}$ given the model parameters, namely, the system noise $\bbalpha$, the observation noise $\sigma^2$, the filter coefficients in $\bbh$, and the initial variance $\sigma_0^2$.
It reads as
\begin{align}
	p(\bby_1,\dots,\bby_T,&\bbx_0,\dots,\bbx_T|\bbh,\bbalpha,\sigma^2,\sigma_0^2)  \\
	&= p(\bbx_0|\sigma_0^2)\prod_{t=1}^{T}p(\bby_t|\bbx_t,\bbh,\sigma^2)p(\bbx_t|\bbx_{t-1},\bbalpha),\nonumber
\end{align}
due to the Markovian property of the state-space model.
From~\eqref{eq:state_space_model}, the transition and observation densities can be written respectively as
\begin{align}
	p(\bbx_t|\bbx_{t-1},\bbalpha) &= \ccalN(\bbx_t|-c\bbL\bbx_{t-1},\bbQ), \\
	p(\bby_t|\bbx_t,\bbh,\sigma^2) &= \ccalN(\bby_t|\bbH\bbx_t,\sigma^2\bbI),
\end{align}
with system noise covariance matrix $\bbQ = \bbB \diag(\bbalpha^2) \bbB^\top$ and graph filter $\bbH$ as in~\eqref{eq:observation_model_linear}.

Maximizing the likelihood, or equivalently minimizing the negative log-likelihood $\ccalL$, w.r.t. the model parameters (graph filter coefficients $\bbh$, edge uncertainties $\bbalpha$, observation noise $\sigma^2$, and initial variance $\sigma_0^2$) is the goal of the learning algorithm to fit the best state space model into the data.
The negative log-likelihood has the closed-form expression
\begin{align}
	\ccalL &= \frac{NT}{2}\log \sigma^2 + \frac{1}{2\sigma^2}\sum_{t=1}^{T}\|\bby_t-\bbH\bbx_t\|_2^2 \nonumber\\
	&+\frac{T}{2}\log|\bbQ| + \frac{1}{2}\sum_{t=1}^{T}(\bbx_t+c\bbL\bbx_{t-1})^\top\bbQ^{-1}(\bbx_t+c\bbL\bbx_{t-1}) \nonumber\\
	&+\frac{N}{2}\log \sigma_0^2 + \frac{1}{2\sigma_0^2}\|\bbx_0\|_2^2.
	\label{eq:likelihood}
\end{align}
Here, the first line is a data fidelity term, whereas the second line measures how well the state variable $\bbx_t$ fits a diffusion process up to an edge-wise uncertainty term $\bbQ$, finally, the last line is the term for the initial state and will be fixed to avoid numerical instability.
We initialize the state as the Normal distribution for simplicity, however, one can use the Riccati solution for more accurate initial conditions~\cite{borovych_data-driven_2022}.
Notice that the likelihood function $\ccalL$ is differentiable and can be minimized via gradient-based methods after inferring the latent state variable $\bbx_t$.
%-------------------------------------------------------------------

%-------------------------------------------------------------------
% EM Algorithm
\subsection{Expectation-Maximization~(EM) Algorithm}

To minimize~\eqref{eq:likelihood}, we consider the EM algorithm, which is a widely-used two-step iterative algorithm for finding the maximum likelihood estimates of parameters in statistical models in the presence of latent variables~\cite{dempster_maximum_1977,baum_statistical_1966}.
The EM algorithm maximizes a lower bound on the likelihood function given any distribution of the latent variables.
Then, it iteratively refines the distribution of the latent variables (E-step) and the model parameters (M-step) until convergence.

% \medskip
\noindent\textbf{E-step:}
Given the model parameters ${\Theta} = \{\bbh, \bbalpha, \sigma^2\}$ and a batch of observations $\{\bby_t\}_{t=1}^T$, the state estimation is the minimum-variance Bayesian inference of the state-space model.
This solution can be derived via Kalman smoothing~\cite{kalman_new_1961}. 
% Note that in an online setting, the Kalman filtering algorithm is used to estimate the state variable $\bbx_t$ given the observations $\bby_t$; however, as we have access to the whole batch of observations, we can use the Kalman smoothing algorithm to estimate the state variable $\bbx_t$ given all the observations $\{\bby_t\}_{t=1}^T$.

% \smallskip
\noindent\textbf{Filtering.}
The Kalman filtering is a closed-form solution of Bayesian inference of states given causal observations.
The algorithm is as follows:

\begin{flalign}
	&\texttt{prediction}
	\left\{
	\begin{aligned}
		&\bbx_t^- = -c\bbL\bbx_{t-1} \\
		&\bbP_t^- = c^2\bbL\bbP^+_{t-1}\bbL^\top + \bbQ
	\end{aligned}
	\right.
	\label{eq:kf_pred}\\
	&\texttt{correction}
	\left\{
	\begin{aligned}
		&\bbK_t = \bbP_t^-\bbH^\top(\bbH\bbP_t^-\bbH^\top + \sigma^2\bbI)^{-1} \\
		&\bbx^+_t = \bbx_t^- + \bbK_t(\bby_t - \bbH\bbx_t^-) \\
		&\bbP^+_t = (\bbI - \bbK_t\bbH)\bbP_t^-
	\end{aligned}
	\right.
	\label{eq:kf_corr}
\end{flalign}
where $\bbK_t \in \reals^{N \times N}$ is the Kalman gain and incorporates the information from a new observation in the raw prediction considering the Bayesian structure of the state-space model.
$\bbP_t = \mathbb{E}[\bbx_t\bbx_t^\top]$ is the covariance matrix of the state variable $\bbx_t$.
For the sake of simplicity, we use $\bbx_t$ instead of $\mathbb{E}[\bbx_t]$ to indicate the mathematical expectation of the state variable.

% \smallskip
\noindent\textbf{Smoothing.}
The Kalman smoothing moves backward over the trajectory and re-corrects the Kalman filter estimations having the information from all the observed data in the batch.
The equations for the Kalman smoother are 
\begin{flalign}
	&\bbJ_{t-1} = -c\bbP^+_{t-1}\bbL(\bbP_t^-)^{-1} \\
	&\hat{\bbx}_{t-1} = \bbx_{t-1} + \bbJ_{t-1}(\hat{\bbx}_t + c\bbL\bbx_{t-1}) \\
	&\hat{\bbP}_{t-1} = \bbP_{t-1}^+ +\bbJ_{t-1}(\hat{\bbP}_t-\bbP_{t-1}^+)\bbJ_{t-1}^\top \\
	&\hat{\bbV}_{t-1} = \bbP_{t-1}^+\bbJ_{t-2}^\top +\bbJ_{t-1}(\hat{\bbV}_t+c\bbL\bbP_{t-1}^+)\bbJ_{t-2}^\top
\end{flalign}
where $\hat{(\cdot)}=\mathbb{E}[(\cdot)|\{\bby_t\}_{t=1}^T]$, $\bbJ_t$ is the smoothing gain helping the backward recursion over mean $\hbx_t$ and covariance $\hbP_t$, and $\hbV_t = \cov[\bbx_t,\bbx_{t-1}|\{\bby_t\}_{t=1}^T]$ is the covariance of the estimated state variable $\hbx_t$ between two consecutive time instants which is required for the computation of the intermediate distribution in the EM algorithm.

Having the state variable $\bbx_t$ and its covariance matrix $\bbP_t$, the intermediate distribution of the likelihood can be computed in the E-step and used in the M-step to update the model parameters as we discuss next.

% \medskip
\noindent\textbf{M-step:}
To apply EM on~\eqref{eq:likelihood}, the expectation of the negative log-likelihood conditioned to observations is required.
This intermediate distribution can be computed as
\begin{align}
	\ccalQ &= \mathbb{E}[\ccalL|\{\bby_t\}_{t=1}^T] \label{eq:intermediate_dist}	\\
	&= \frac{NT}{2}\log\sigma^2 + \frac{T}{2}\log|\bbQ| \nonumber\\
	&+ \frac{1}{2\sigma^2}\tr
	\left(
	\sum_{t=1}^{T}\bby_t\bby_t^\top - \bbH\hat{\bbx}_t\bby_t^\top - \bby_t\hat{\bbx}_t^\top\bbH^\top + \bbH\hat{\bbP}_t\bbH^\top
	\right) \nonumber\\
	&+\frac{1}{2}\tr
	\left(
		\bbQ^{-1}\sum_{t=1}^{T}\hat{\bbP}_t+c\bbL\hat{\bbV}_t+c\hat{\bbV}_t\bbL^\top + c^2\bbL\hat{\bbP}_{t-1}\bbL^\top
	\right)
	\nonumber
	. 
\end{align}
To minimize $\ccalQ$ w.r.t. the model parameters, one can compute the gradients of~\eqref{eq:intermediate_dist}.
% These gradients can be calculated in closed-form via
% \begin{align*}
% 	\frac{\partial\ccalQ}{\partial\bbH}
% 	&= \frac{1}{\sigma^2} \left(\sum_{t=1}^{T}\hat{\bbx}_t\bby_t^\top - \bbH\hat{\bbP}_t \right), \\
% 	\frac{\partial\ccalQ}{\partial\sigma^{-2}}
% 	&=\frac{NT\sigma^2}{2} -\frac{1}{2}\tr\left(\sum_{t=1}^{T} \bby_t\bby_t^\top - 2\bbH\hat{\bbx}_t\bby_t^\top + \bbH\hat{\bbP}_t\bbH^\top\right), \\
% 	\frac{\partial\ccalQ}{\partial c}
% 	&=\tr\left(\bbQ^{-1}\bbL\sum_{t=1}^{T}\hat{\bbV}_t\right)
% 	+c~\tr\left(\bbQ^{-1}\bbL\sum_{t=1}^{T}\hat{\bbP}_{t-1}\bbL\right), \\
% 	\frac{\partial\ccalQ}{\partial\bbQ^{-1}}
% 	&=\frac{T}{2}\bbQ-\frac{1}{2}\sum_{t=1}^{T}\hat{\bbP}_t+c{\bbL}\hat{\bbV}_t,
% \end{align*}
% which can be further generalized to the model parameters with the chain rule.
The model parameters can be obtained via closed-form solution of the gradients or be updated using gradient-based optimization methods.

%-------------------------------------------------------------------
The EM algorithm iterates between the E-step, and the M-step until convergence.
The computational complexity of the E-step in each iteration is of order $\ccalO(TN^3)$ due to the inverse matrix operation computation of the Kalman gain in~\eqref{eq:kf_corr} where $N$ is the number of nodes in the graph.
The M-step is computationally less expensive, and it is governed by $N$-by-$N$ matrix-matrix products in each timestamp, hence, the computations can be done in $\ccalO(TN^2)$.
% In conclusion, this algorithm does not scale well with the number of nodes in the graph, but it has linear complexity with respect to the number of observations $T$ thanks to the recursive structure of the Kalman filter and smoother.
%-------------------------------------------------------------------
%%%---------------------------------------------------------------------

%%%---------------------------------------------------------------------
% \section{Model-based Inference and Learning}
% \label{sec_unrolling}
% \input{Sections/unrolling}
%%%---------------------------------------------------------------------

%%%---------------------------------------------------------------------
\section{Model-based Inference and Learning}
\label{sec_rgkn}
\begin{figure*}[ht]
	\centering
	% !TEX root = ../main.tex

\def \thisplotscale {1.77}
\def \unit {\thisplotscale cm}

\tikzstyle {Phi} = [rectangle, 
thin,
minimum width = 1.15*\unit, 
minimum height = \sumshift*\unit, 
anchor = center,
draw,
fill = cyan!20]

\tikzstyle {sum} = [circle, 
thin,
minimum width  = 0.3*\unit, 
minimum height = 0.3*\unit, 
anchor = center,
draw,
fill = cyan!20]

\def \deltax {1.6}
\def \deltay {1.2}
\def \deltagat {1.2}
\def \sumshift {0.4}

{\footnotesize\begin{tikzpicture}[x = 1*\unit, y = 1*\unit]
		
		\pgfdeclarelayer{bg}    % declare background layer
		\pgfsetlayers{bg,main}  % set the order of the layers (main is the standard layer)

		% Begin by drawing an empty node for initializing the chain
		\node (first) [] {};    
		
		% nodes
		\path
		% (first) ++ (.75*\deltax,0) node (enc) [trapezium, draw, rotate=-90, fill = mygreen!30, minimum height= .5*\unit]{\rotatebox{-180}{Encoder}}
		(first) ++ (.75*\deltax,0) node (enc) [Phi, fill=mygreen!30]{Encoder}
		++ (3*\deltax, 1*\deltay) node (pred) [Phi]
		{$\begin{aligned}
			&\text{Prediction sub-module:} \\
			&\bbx_t^- = -c\bbL\bbx_{t-1} \\
			&\bbP_t^- = c^2\bbL\bbP_{t-1}^+\bbL^\top + \bbB\diag(\bbalpha)\bbB^\top 
		\end{aligned}$}
		++ (0,-1*\deltay) node (upd) [Phi] 
		{$\begin{aligned}
			&\text{Correction sub-module:} \\
			&\bbx_t = \bbx_t^- + \bbK_t(\hbx_t - \bbx_t^-) \\
			&\bbP_t = \bbP_t^- - \bbP_t^-\bbK_t 
		\end{aligned}$}
		% (upd.east) ++ (.75*\deltax, .5*\deltay) node (dec) [trapezium, draw, rotate=90, fill = mygreen!30, minimum height= .5*\unit]{Decoder}
		(upd.east) ++ (1.25*\deltax, .5*\deltay) node (dec) [Phi, fill = mygreen!30]{Decoder}
		(upd.west) ++ (-.25*\deltax,0) node(c1){}
		(dec.west) ++ (-.35*\deltax,0) node(c2){}
		(c2) ++ (0,.5*\deltay) node(c3){}
		(enc.center) ++ (.6*\deltax,.5*\deltay) node(net)[Phi]{Inference}
		(net) ++ (0,.5*\deltay) node(infer)[Phi]{RNN}
		(infer.east) ++ (.25*\deltax,0) node (h1) []{}
		(infer.north) ++ (0,.25*\deltay) node (h2)[]{}
		(enc.center)	++ (0,1*\deltay) node (c4) {}
		(enc.west) ++ (0,1.5*\deltay) node (b1) {}
		(enc.east) ++ (1.4*\deltax,.3*\deltay)	node (b2) {}
		(b1) ++ (2.35*\deltax,0) node (b3) {}
		(upd.south)	++ (1.15*\deltax,-.05*\deltay) node (b4) {};

		%edges
		\path[-stealth]
			(first)	edge[above]	node{$\bby_t$}	(enc)
			(c1.center)	edge[above] node{$\hbx_t$} (upd.west)
			(c2.center)	edge[]	node[above]{$\bbx_{t}$} (dec)
			(c3.center)	edge[]	node[above]{$\bbx_{t}$} node[below]{$\bbP_t^+$}	(pred.east)
			(pred.south)	edge[]	node[left]{$\bbx_t^-$} node[right]{$\bbP_t^-$}	(upd.north)
			(dec.east)	edge[above]	node{$\hby_t$}	++ (.3*\deltax,0)	node{}
			(h2.center) edge[left]	node{$\bbh_t$}	(infer.north)
			(c4.center)	edge[above]	node{$\bbsigma_t$}	(infer.west)
			(infer.south) edge[]	node{}	(net.north);

		\path[draw]
			(enc)		-|	node{} (c1.center)
			(upd.east)	-|	node{}	(c2.center)
			(c2.center)	--	node{}	(c3.center)
			(infer.east) -- (h1.center)
			(h1.center) |- (h2.center)
			(enc.north) -| (c4.center);

		% Draw box around layers
		\begin{pgfonlayer}{bg}
			\node[fit=(b3)(b4)(c3)(c1)(pred.north)(upd.south), draw,inner sep = 0*\deltax,fill = red!30, fill opacity = 0.5,draw = white, rounded corners=10](l1) {};

			\node[fit=(b1)(b2)(infer)(net)(h2)(h1), draw,inner sep = 0*\deltax,fill = red!30, fill opacity = 0.5,draw = white, rounded corners=10](l2) {};
			
			% \draw[-stealth,dashed] (l2.south) -- node[right]{$\bbh_t = \{c,\bbalpha,\bbK_t\}$} (l2.south |- l1.north);
		\end{pgfonlayer}
		
		\path[-stealth]
			(net.east) edge[above]	node{$\{c,\bbalpha_t,\bbK_t\}$}	(net.east -| l1.west);

		\path (l1.north) ++ (0,+.001*\deltay) node (l3) [draw, fill=mygreen!30]{Kalman Module};
		\path (l2.north) ++ (0,+.001*\deltay) node (l4) [draw, fill=mygreen!30]{Inference Module};

\end{tikzpicture}}
	\caption{Recurrent Graph Kalman Network block diagram.
	Encoder: takes the observations as input and transfer them into the latent domain.
	The Kalman module: updates the data in the state space assuming a diffusion process over the graph.
	Inference module: estimates the parameters of the Kalman module given the observations indirectly and a hidden recurrent layer.
	Decoder: receives the output of Kalman module and transfers it back to the observation domain.}
\end{figure*}
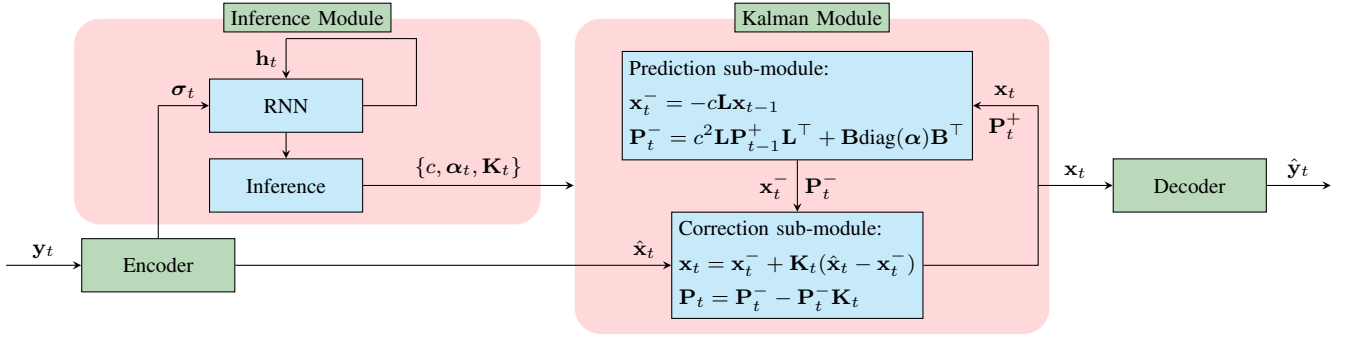

The proposed Kalman-based approach is mathematically tractable and optimal, but it has three main limitations: (i) low expressivity due to the linearity of the model; (ii) strict mathematical assumptions on the data; and (iii) high computational complexity in the number of nodes.
To address these issues, we propose a novel graph Kalman network (GKNet) which is a model-based deep inference and learning framework for data on graphs.
GKNet consists of four modules.
First, there are the encoder and the decoder modules which are graph convolutional neural networks (GCNNs)~\cite{isufi_graphfilters_2024} that transform the observations into the state domain and vice-versa with a nonlinear mapping to increase the model expressivity.
Second, there are prediction and correction modules, forming the Kalman module, which are graph filters that account for the prediction and correction steps in Kalman filter and smoother, respectively. Parametrizing the Kalman gain as a graph filter in the Kalman module relaxes the statistical assumptions of the model and expands its applicability.
Finally, there is a recurrent module as an inference model estimating the parameters of the prediction and correction modules that reduces the number of parameters and recursions of the model drastically.

%-------------------------------------------------------------------
\subsection{Encoder and Decoder}
The observation equation~\eqref{eq:observation_model_linear} connects the observations $\bby_t$ to the state variable $\bbx_t$ via a linear graph filter $\bbH$.
The encoder and decoder modules are responsible for transforming the data into the latent space and vice-versa similar to the graph filter $\bbH$.
To preserve the graph filter principle but improve its expressivity, we consider graph convolutional neural networks (GCNNs) as the encoder and decoder modules.
A GCNN layer can be defined as
\begin{equation}
	\tbx_t^{(l+1)} = \text{ReLU}\left(\sum_{k=0}^{K} h_{kl} \bbL^k \tbx_t^{(l)}\right),\quad l=0,\dots,L-1
\end{equation}
where the trainable parameters $\bbh_l = [h_{0l},\dots,h_{Kl}]^\top$ are the coefficients of the graph filter in each layer $l$, $\text{ReLU}(\cdot)$ is the chosen non-linear activation function, and $\tbx_t^{(l)}$ is the feature vector of the $l$-th layer~\cite{gama_convolution_2020}.

The encoder transforms the observations $\bby_t$ into the state domain $\bbx_t$ via a graph convolutional neural network.
This module receives the observation $\bbx^{(0)}_t = \bby_t \in \reals^{N}$ (the input is zero-padded at nodes with missing value) and returns $\tbx_t = \tbx_t^{(L)} \in \reals^{N}$ as a rough estimation of the state mean and $\bbsigma_t \in \reals^N$ as the estimation of observation noise.
Hence, the encoder function $\Phi_{\text{enc}}(\cdot): \reals^{N} \rightarrow \reals^{N \times 2}$ is defined as
\begin{equation}
	[\tbx_t, \bbsigma_t] = \Phi_{\text{enc}}(\bby_t,\bbL;\ccalH_{\text{enc}})
\end{equation}
where $\Phi_{\text{enc}}(\cdot)$ is a GCNN with $L$ layers and $F$ features per layer and $\ccalH_{\text{enc}} = [\bbh_l]_{l=0}^L$ contains the learnable parameters of the GCNN.
% Using this non-linear transformation, the model can capture more complex patterns in the data to explore a latent space that diffuses linearly over the graph.

The decoder is also a GCNN followed by a readout layer based on the application and desired targets.
The decoder's input is the processed latent state $\bbx_t$ and the output is the estimation of desired targets $\hby_t$.
The decoder is defined as
\begin{equation}
	\hby_t= \Phi_{\text{dec}}(\bbx_t,\bbL;\ccalH_{\text{dec}})
\end{equation}
where $\ccalH_{\text{dec}}$ contains the learnable parameters of $\Phi_\text{dec}(\cdot)$, $\bbx_t^{(0)} = \bbx_t$ is the input, and $\hby_t = \bbx_t^{(L)}$ is the output.

%-------------------------------------------------------------------
% \vspace{-.5cm}
\subsection{Kalman Module}
The Kalman module accounts for the statistical computations done via the Kalman filter inside the state space.
The module consists of two main parts: prediction and update.
The prediction sub-module estimates the state variable's statistics based on the previous state and the edge uncertainty vector $\bbalpha$ while the update sub-module corrects this estimation based on the observation%
\footnote{For the sake of simplicity, the Kalman smoothing is removed in the presented neural-aided model.
However, it can be added explicitly in the Kalman module where the matrix $\bbJ_t$ is a graph filter with parameters derived from the inference module.
Implicitly, the reverse recursions can be  accounted for the inference module by using a bi-directional recurrent network.}%
.

%-------------------------------------------------------------------
% \medskip
\noindent\textbf{Prediction sub-module.}
This module accounts for the prediction step in Kalman filter in~\eqref{eq:kf_pred}.
Given the diffusivity factor $c$ and edge uncertainties $\bbalpha$, this module predicts the mean $\bbx_t^-$ and covariance matrix $\bbP_t^-$ of the state variable based on Equation~\eqref{eq:kf_pred}.
The initial state $(\bbx_{t-1},\bbP_{t-1})$ is either provided by the previous recurrent layer or initialized by $(\bbx_0,\bbP_0) = (\bbzero, \bbI)$.
Thus, this module can be defined as
\begin{equation}
	(\bbx_t^-,\bbP_t^-) = \Psi_{\text{pred}}((\bbx_{t-1},\bbP_{t-1}),\bbL;c,\bbalpha)
\end{equation}
where $\Psi_{\text{pred}}(\cdot)$ recurrently predicts the state variable based on Kalman equations~\eqref{eq:kf_pred}.

%-------------------------------------------------------------------
% \medskip
\noindent\textbf{Correction sub-module.}
This accounts for the update steps of Kalman filter in~\eqref{eq:kf_corr}.
We replace the Kalman gain $\bbK_t$ with a graph filter $\hbK_t(\bbL,\sigma^2)$ and define the correction module as
\begin{align}
	\bbx_t &= \bbx_t^-  + \hbK_t (\hbx_t - \bbx_t^-) \\
	\bbP_t &= (\bbI - \hbK_t)\bbP_t^-
\end{align}
where $\bbx_t^-$ and $\bbP_t^-$ are the predicted mean and covariance matrix of the state variable from the prediction module, $\hbx_t$ is the output of the encoder, and $\hbK_t$ is a graph filter approximating the Kalman gain $\bbK_t$.
This graph filter aims to mimic the Kalman gain $\bbK_t$ in a scalable and data efficient manner with fewer parameters.

To justify the module, assume the encoder has estimated the graph filter $\bbH$ well and  $\bby_t = \bbH\hbx_t$ holds. Then we can write
\begin{equation}
	\bbx_t = \bbx_t^- + \bbP_t^- \bbH^\top(\bbH \bbP_t^-\bbH^\top + {\sigma}^2\bbI)^{-1}\bbH(\hbx_t - \bbx_t^-).
\end{equation}
As the graph filter $\bbH$ is a function of $\bbL$, the update equation can be approximated with a graph filter up to the Eigenspace mismatch between $\bbL$ and $\bbP_t^-$ as
\begin{equation}
	\bbx_t = \bbx_t^- + \bbP_t^- \hbK_t(\bbL,\sigma^2)(\hbx_t - \bbx_t^-)
\end{equation}
where $\hbK_t(\cdot)$ is a function of the graph Laplacian and the observation noise variance.
We consider a graph filter as $\hbK_t(\bbL,\sigma^2) = \sum_{k=0}^{K} h_{kt}(\sigma^2) \bbL^k$ where the parameters $h_{kt}(\sigma^2)$ are dependent on the estimated observation noise variance.
Similarly, the update equation for the covariance matrix can be rewritten as
\begin{equation}
	\bbP_t^+ = (\bbI - \bbP_t^-\hbK_t(\bbL,\sigma^2))\bbP_t^-.
\end{equation}

% If more expressivity is needed, we can use a GCNN in place of the graph filter $\hbK_t(\cdot)$.
%-------------------------------------------------------------------
\subsection{Inference Module}

To estimate the parameters in the predictions and correction modules, one can unroll the Kalman equations and learn the parameters in an end-to-end manner.
However, this limits the model to a fixed temporal window $T$, high computational complexity, and linear time-invariant systems.
The inference module encodes the parameters of the prediction and correction modules into the state of a recurrent neural network (RNN) that is responsible to capture temporal behavior of the state variable through its model parameters.
The recurrent unit is defined as
\begin{align}
	&\bbz_t = \text{ReLU}(\text{BN}(\bbU_{in}\bbsigma_t) + \bbU_z\bbh_{t-1}) \nonumber \\
	&\hbh_t = \text{ReLU}(\text{BN}(\bbU_{out}\bbsigma_t) + \bbU_h\bbh_{t-1}) \nonumber \\
	&\bbh_t = \bbz_t \odot \bbh_{t-1} + (1-\bbz_t) \odot \hbh_t
	\label{eq:gru}
\end{align}
where $\sigma(\cdot)$ is the sigmoid activation function, $\text{BN}(\cdot)$ is the batch normalization, $\odot$ is the element-wise multiplication,
and $\bbU_{in} \in \reals^{\tilde{K} \times N }$, $\bbU_{out} \in \reals^{\tilde{K} \times N }$, $\bbU_z \in \reals^{\tilde{K} \times \tilde{K}}$, and $\bbU_h \in \reals^{\tilde{K} \times \tilde{K}}$ are the learnable parameters of the RNN where $\tilde{K}$ is the number of parameters in Kalman module.

The Kalman gain filter $\hbK_t$ is completely defined via its coefficients $\bbh_t^{(0)} = [h_{0t}^{(0)},\dots,h_{Kt}^{(0)}]^\top \in \reals^{K+1}$, and the diffusivity factor $c\in \reals$ is a scalar.
However, the edge uncertainties $\bbalpha \in \reals^M$ increase with graph size and can be limiting for model scalability.
To address this issue, we consider the edge uncertainties $\bbalpha$ to be the output of an innovation process with an edge convolutional filter $\bbH_e(\bbL_1)$ defined as
\begin{equation}
	\bbalpha_t = \sum_{k=0}^{K} h_{kt} (\bbB^\top \bbB)^k \bbw_t = \bbH_e(\bbL_1)\bbw_t,
\end{equation}
where $\bbH_e(\bbL_1) = \sum_{k=0}^{K} h_{kt} (\bbB^\top \bbB)^k $ is an edge convolutional filter of order $K$ operating on edge Laplacian $\bbL_1 = \bbB^\top\bbB$, $\bbh_t = [h_{0t},\dots,h_{K_1 t}]^\top$ are trainable filter coefficients, and $\bbw_t$ is the innovation process input.
The input signal $\bbw_t$ can be any known signal and holds prior knowledge about the edge uncertainties.
For example, $\bbw_t = \bbB^\top \bbd$, where $\bbd$ is the degree vector of the graph, is a prior enforcing the edges connected to the nodes with higher degrees more susceptible to uncertainty which is tied to relative perturbations of graphs.
Note that in the GKNet, the edge uncertainties $\bbalpha_t$ are varying over time, unlike the EM-based approach which needed them static over time for the time invariance of the system.

In summary, the inference module computes the parameters required in the Kalman module $\bbh_t = [c_t, \bbh_t^{(0)\top},  \bbh_t^{\top}]^\top \in \reals^{2K+3}$ through the state of the RNN in~\eqref{eq:gru}.
The size of the RNN state $\bbh_t$ is independent of the size of the graph as the number of filter coefficients in the graph filters is fixed.
However, the size of filters $\bbU_{in}$ and $\bbU_{out}$ are dependent on the number of nodes in the graph, as they are incorporating the observation statistics effects, indirectly computed in $\bbsigma_t$, into the model parameters $\bbh_t$.
Although this operation is still computationally efficient, it confines the model transferability.
To tackle this one can remove this dependency, use a global aggregation of the observation statistics, or fine-tune the model on the target graph domain.
The computational comlexity of the inference module is $\ccalO(NT(2K+3))$ due to the matrix-vector multiplications in the RNN, the encoder and decoder modules are $\ccalO(MKFL)$, and the Kalman module is $\ccalO(MNK)$.
Hence, the computational cost of GKNet reduced significantly compared to the EM approach which is $\ccalO(TN^2)$.

%-------------------------------------------------------------------
% \vspace{-.7cm}
\subsection{Training}
The loss function is inspired by the likelihood function~\eqref{eq:likelihood} and enforces the state to follow an evolution over the graph.
The loss function reads as
\begin{equation}
	\ccalL_t = \|\hby_t - \bby_t\|_2^2 + 
	\lambda \|\bbx_t+c\bbL\bbx_{t-1}\|_{\bbQ^{-1}}^2
	\label{eq:loss}
\end{equation} 
where $\lambda$ is the regularization weight and $\|\cdot\|_\bbA^2 = (\cdot)^\top \bbA (\cdot)$ is the quadratic form over the matrix $\bbA$.
The first term is the data fidelity term and the second term is the smoothness term enforcing the state to follow a diffusion process over the graph up to the edge uncertainty measure $\bbQ^{-1}$.
The inverse operation is computationally cheap as the matrix $\bbQ = \bbB\diag(\bbalpha)\bbB^\top$ only changes via the diagonal edge uncertainty term $\bbalpha$ and the rest can be pre-computed.
It means we can write the inverse of $\bbQ^{-1} = \bbB^{\dagger_L}\diag(\bbalpha^{-1})\bbP^{\dagger_R}$ and pre-compute the left and right psuedo-inverses of $\bbB$ as $\bbB^{\dagger_L}$ and $\bbB^{\dagger_R}$, respectively~\cite{bapat_graphs_2010}.
%-------------------------------------------------------------------
%%%---------------------------------------------------------------------

%%%---------------------------------------------------------------------
\section{Numerical Results}
\label{sec_numeric}
We evaluate GKNet in different scenarios compared with state-of-the-art and baseline alternatives.
In all experiments, the ADAM optimizer is used to train the model.
The learning rate is set to $10^{-3}$ with a batch size of 32.
The training is performed for 100 epochs with early stopping based on the validation loss.
The reported results are the average of 20 runs with different random seeds.
The standard variations are omitted as they are all the same order of magnitude $\ccalO(10^{-3})$.
Our experiments aim to provide insights into the model components by answering the following three main research questions:
\begin{itemize}[itemindent = 20pt]
\item[\textbf{RQ.1}:] How does the proposed model perform in high data regime for interpolation and extrapolation tasks when compared with alternatives?
\item[\textbf{RQ.2}:] Can the proposed model generalize well in a low data regime for interpolation and extrapolation tasks?
\item[\textbf{RQ.3}:] Can the GKNet perform well in applications with missing values? 
\item[\textbf{RQ.4}:] How do the different components of the proposed model affect the performance such as the regularization term and uncertainty model?
\end{itemize}
We answer these questions in three tasks: multivariate time-series forecasting; spatiotemporal signal interpolation;  and tracking dynamic systems.
For further details, the codes are available here.
% \footnote{https://github.com/mohammad-sabbaqi-95/GKNet}

%-------------------------------------------------------------------
\subsection{Multivariate Time-Series Forecasting}

We consider a traffic forecasting application on two different benchmark datasets, namely, METR-LA and PEMS-BAY.
METR-LA contains four months of recorded traffic data over 207 nodes on the highways of Los Angeles County with 5-minute resolution~\cite{li_diffusion_2017}.
PEMS-BAY includes six months of traffic information over 325 nodes in the Bay Area with the same resolution~\cite{li_diffusion_2017}.
The graph is constructed by applying a Gaussian threshold kernel over the road network distance matrix as in~\cite{li_diffusion_2017}.
The goal is to predict the traffic load in different time horizons $h =\{3, 6, 12\}$ having the time series for the last $T = \{3, 6, 12\}$ time steps.

Both datasets are divided chronologically for training and testing.
Different percentages of the data are assigned for the training of the model in the different experiments.
The encoder and decoder are GCNNs with two layers and third-order filters operating on the normalized Laplacian matrix of the graph.
The number of features in the middle layer is $F = 16$.
The regularization weight in~\eqref{eq:loss} is $\lambda = 0.05$.
The evaluation metric is the normalized root mean square error (RMSE).

% \medskip
\noindent\textbf{Baselines.}
A combination of statistical and data-driven models are used with the model.
The statistical models used to address forecasting problems are:
\begin{itemize}
	\item v-ARIMA: vector autoregressive integrated moving average model using Kalman filter~\cite{box_time_2015}.
	\item G-VARMA: graph vector autoregressive moving average model~\cite{isufi_forecasting_2019}.
	\item GP-VAR: graph polynomial autoregressive model~\cite{isufi_forecasting_2019}.
	\item Grad-based: proposed model solving the maximum likelihood~\eqref{eq:likelihood} via gradient descent.
	\item EM-based: proposed model solving the maximum likelihood via expectation maximization. 
\end{itemize}

\noindent
The data-driven models are:
\begin{itemize}
	\item Graph WaveNet: A hybrid convolutional model for time series over graphs~\cite{wu_graph_2019}.
	\item STGCN: a spatial temporal graph convolutional network~\cite{yan_spatial_2018}.
	\item GGRNN: graph gated recurrent neural network~\cite{ruiz_gated_2020}.
	\item DCRNN: diffusion convolutional recurrent neural network~\cite{li_diffusion_2017}.
	\item GTCNN: a product graph based graph-time convolutional neural network \cite{sabbaqi_gtcnn}.
\end{itemize}

% experiment #1
% \medskip\noindent\green{
% \framebox[\columnwidth]{\parbox{.9\linewidth}{Experiment \#1}}}\medskip

% \medskip
\noindent\textbf{Performance in high data regime.}
75\% of the data is used for the training and validation.
As shown in Table~\ref{table:traffic_high}, data-driven models outperform statistical models.
Even though the proposed GKNet model is designed to handle uncertainty and learn from fewer data samples, it is still performing competitively with the state-of-the-art models as it enjoys the expressive power of GCNNs in the encoder and decoder modules.
Among the statistical models, the proposed model outperforms the rest, and it manages to keep up the performance in longer prediction horizons.
This generalization power is a direct result of estimating the uncertainty in the evolution of the data.

% table for high data regime traffic forecasting
\begin{table*}[t]
	\centering
	\caption{Multivariate time series forecasting performance for traffic prediction on METR-LA and PEMS-BAY datasets. The experiments are performed in a high data regime where 75\% of the data is used for the training and validation of the models.}
	\label{table:traffic_high}
	% \medskip
	\newcolumntype{C}{>{\centering\let\newline\\\arraybackslash\hspace{0pt}}m{1cm}}

\renewcommand{\arraystretch}{1.25}
\begin{tabular}{m{2.75cm}CCC|CCC}
	\hline\hline
	\multirow{2}{1cm}{\centering nRMSE}
	&\multicolumn{3}{c}{METR-LA}
	&\multicolumn{3}{c}{PEMS-BAY}
	\\ \cline{2-7}
	&$h=3$	&$h=6$	&$h=12$
	&$h=3$	&$h=6$	&$h=12$
	\\ \hline
	ARIMA
	&0.3511	&0.3630	&0.3796
	&0.3768	&0.3924	&0.3904
	\\
	G-VARMA~\cite{isufi_forecasting_2019}
	&0.3478	&0.3692	&0.4022
	&0.3266	&\blue{0.3384}	&\blue{0.3572}
	\\
	GP-VAR~\cite{isufi_forecasting_2019}
	&0.3461	&0.3319	&0.4278
	&0.3583	&0.3844	&0.4206
	\\
	Grad-based (ours)
	&\blue{0.3114}	&\blue{0.3201}	&\blue{0.3249}
	&0.3615	&0.3658	&0.3897
	\\
	EM-based (ours)
	&0.3595	&0.3606	&0.3713
	&\blue{0.3070}	&0.3404	&0.3585
	\\
	\noalign{\vskip 2pt}\cdashline{1-7}[.4pt/1pt]\noalign{\vskip 2pt}
	Graph WaveNet~\cite{wu_graph_2019}
	&0.2621	&0.2645	&0.2800	
	&0.2391	&0.2437	&0.2924
	\\
	STGCN~\cite{yan_spatial_2018}
	&0.2355	&0.2820	&0.2913	
	&\green{0.2187}	&\green{0.2223}	&0.2889
	\\
	GGRNN~\cite{ruiz_gated_2020}
	&\green{0.2095}	&0.2489	&0.2854	
	&0.2429	&0.2655	&0.2920
	\\
	DCRNN~\cite{li_diffusion_2017}
	&0.2508	&0.2580	&0.2666	
	&0.2391	&0.2489	&\green{0.2617}
	\\
	GTCNN~\cite{sabbaqi_gtcnn}
	&0.2296	&0.2486	&\green{0.2498}	
	&0.2379	&0.2708	&0.2833
	\\
	GKNet (ours)
	&0.2393	&\green{0.2400}	&0.2653	
	&0.2488	&0.2559	&0.2950
	\\[2pt]
	\hline\hline
\end{tabular}
\renewcommand{\arraystretch}{1}
\end{table*}

% experiment #2
% \medskip\noindent\green{
% \framebox[\columnwidth]{\parbox{.9\linewidth}{Experiment \#2}}}\medskip

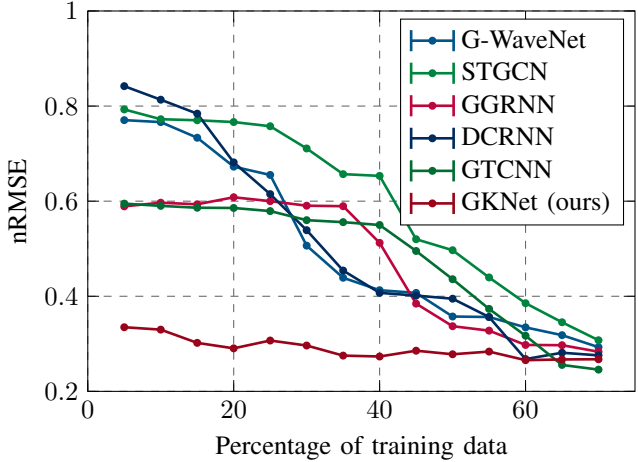
\begin{figure}[t]
	\centering
	\begin{tikzpicture}
\begin{axis}[
width=\linewidth,
height=0.75\linewidth,
axis line style={black},
legend cell align={left},
legend style={draw=black},
tick align=outside,
x grid style={dashed,black!60},
xlabel={Percentage of training data},
xmajorticks=true,
xmin=0, xmax=75,
xtick style={color=black},
y grid style={dashed,black!60},
ylabel={nRMSE},
ymajorticks=true,
ymin=0.2, ymax=1,
ytick style={black},
xtick align=inside,
ytick align=inside,
grid = both,
legend image post style={sharp plot,|-|}
]
\addplot [mark =*,line width=1pt,pennlighterblue, mark size=1pt,error bars/.cd,
        y dir=both, y fixed=0.001]
table {%
5     0.770432
10    0.766313
15    0.733430
20    0.672599
25    0.655175
30    0.506472
35    0.438884
40    0.412760
45    0.407060
50    0.357396
55    0.356375
60    0.334572
65    0.3182004
70    0.292957
};
\addlegendentry{G-WaveNet}

\addplot [mark =*,line width=1pt,pennlightergreen, mark size=1pt,error bars/.cd,
        y dir=both, y fixed=0.001]
table {%
5     0.792968
10    0.772364
15    0.770169
20    0.766414
25    0.757613
30    0.710850
35    0.656834
40    0.653189
45    0.519984
50    0.496847
55    0.439612
60    0.385387
65    0.345346
70    0.307478
};
\addlegendentry{STGCN}

\addplot [mark=*,line width=1pt,pennlighterred, mark size=1pt,error bars/.cd,
        y dir=both, y fixed=0.001]
table {%
5     0.589093
10    0.596742
15    0.593427
20    0.608195
25    0.600051
30    0.590477
35    0.589441
40    0.512348
45    0.384547
50    0.337037
55    0.327717
60    0.297634
65    0.297071
70    0.2833626
};
\addlegendentry{GGRNN}

\addplot [mark=*,line width=1pt,pennblue, mark size=1pt,error bars/.cd,
        y dir=both, y fixed=0.001]
table {%
5     0.841900
10    0.813411
15    0.784036
20    0.681583
25    0.614845
30    0.538959
35    0.454101
40    0.407186
45    0.401417
50    0.394856
55    0.356351
60    0.267928
65    0.281242
70    0.275708
};
\addlegendentry{DCRNN}

\addplot [mark=*,line width=1pt,penngreen, mark size=1pt,error bars/.cd,
        y dir=both, y fixed=0.001]
table {%
5     0.5948169
10    0.5901044
15    0.5861925
20    0.5856974
25    0.5791732
30    0.5600611
35    0.5558552
40    0.5499216
45    0.4952552
50    0.4358522
55    0.3734541
60    0.3169792
65    0.2555597
70    0.245745
};
\addlegendentry{GTCNN}

\addplot [mark=*,line width=1pt,pennred, mark size=1pt,error bars/.cd,
        y dir=both, y fixed=0.001]
table {%
5     0.334870
10    0.329823
15    0.301999
20    0.290487
25    0.306917
30    0.296406
35    0.275203
40    0.273318
45    0.285511
50    0.278007
55    0.283662
60    0.265606
65    0.267130
70    0.267542
};
\addlegendentry{GKNet (ours)}

\end{axis}

\end{tikzpicture}
	\caption{The performance of data-driven models on traffic forecasting task in a low data regime. The experiments are conducted on METR-LA dataset for $h = 6$ prediction horizon.}
	\label{fig:traffic_low}
\end{figure}

% \medskip
\noindent\textbf{Performance in low data regime.}
5\% of the data is used for the training and validation of the models.
These experiments are conducted on the METR-LA dataset for $h = 6$ prediction horizon having different amounts of training samples\footnote{The same trend can be seen in all the different experiment settings, however, we report only for $h = 6$ to preserve space.}.
Fig.~\ref{fig:traffic_low} indicates that the proposed model performs decently even in a low data regime as it transforms the available data into a domain with simple dynamic behavior.
The performance of the other data-driven models drops drastically as the amount of training data decreases.
This is due to the fact that these models either diverge in the training phase as few data samples are available or their parameters need to be reduced to a point where they are not expressive enough for the data.

% figure for low data regime traffic forecasting

\subsection{Multivariate Time-Series Interpolation}

% task definiton

The proposed model is evaluated on the task of multivariate time series interpolation.
Assume a window of time series data $\bbY \in \reals^{N \times T}$ with partial observations $\bbY_{\text{obs}} = \bbM \odot \bbY$ where $\bbM \in \{0,1\}^{N \times T}$ is the mask matrix.
The goal is to estimate the missing values in $\bbY$ given the observations $\bbY_{\text{obs}}$.

The experiments are conducted on two benchmark weather datasets, namely, NOAA, and Molene.
The NOAA dataset contains weather information over 109 stations in the US with a year of hourly measurements.
The Molene dataset includes hourly weather data over 32 stations in France for a month of data.
The same setting as~\cite{isufi_forecasting_2019} is used to construct the graph.
A portion of the data is randomly selected as the observed data and the rest is used for the evaluation of the model.

The encoder and decoder are GCNNs with two layers and second-order filters operating on the normalized Laplacian matrix of the graph.
The number of features in the middle layer is $F = 8$.
The regularization weight in~\eqref{eq:loss} is $\lambda = 0.15$.
% The evaluation metric is the normalized root mean square error (nRMSE).

% \medskip
\noindent
\textbf{Performance in high data regime.}
75\% of the data is used for the training and validation of the models while more than 70\% of the nodes are observed.
As the results in Table~\ref{table:weather_high} indicate, the EM-based proposed algorithm outperforms the other statistical baselines due to the flexibility of the model and the ability of the EM algorithm to handle incomplete data.
Among the data-driven models, the proposed model performs better than the alternatives on the Molene dataset as it is a relatively small dataset and the proposed model is designed for such situations.
However, the performance of the proposed model is not as good as the other data-driven models on the NOAA dataset where more training samples are available.
The performance of the model remains almost constant as the amount of observed data decreases which can be interpreted as having less information about the data.

% table for high data regime weather interpolation
\begin{table*}[t]
	\centering
	\medskip
	\caption{Multivariate time series interpolation performance for weather temperature on Molene and NOAA datasets. The experiments are performed for different node observation ratios~(90,80,70\%) in a high data regime where 75\% of the data is used for the training and validation.}
	\label{table:weather_high}
	% \medskip
	\newcolumntype{C}{>{\centering\let\newline\\\arraybackslash\hspace{0pt}}m{1.25cm}}

\renewcommand{\arraystretch}{1.25}
\begin{tabular}{m{2.75cm}CCC|CCC}
	\hline\hline
	\multirow{2}{1cm}{\centering nRMSE}
	&\multicolumn{3}{c}{Molene}
	&\multicolumn{3}{c}{NOAA}
	\\ \cline{2-7}
	&$90\%$	&$80\%$	&$70\%$
	&$90\%$	&$80\%$	&$70\%$
	\\ \hline
	ARIMA
	&0.3344	&0.3701	&0.3802
	&0.3391	&0.3493	&0.3741
	\\
	G-VARMA~\cite{isufi_forecasting_2019}
    &0.2524	&0.2887	&0.2915
	&0.2767	&0.3633	&0.3772
	\\
	GP-VAR~\cite{isufi_forecasting_2019}
	&0.2673	&0.2741	&0.2747
	&0.3375	&0.3761	&0.3584
	\\
	Grad-based (ours)
	&0.2787	&0.2958	&0.3491
	&0.2840	&0.3106	&0.2938
	\\
	EM-based (ours)
	&\blue{0.2335}	&\blue{0.2578}	&\blue{0.2608}
	&\blue{0.2260}	&\blue{0.2677}	&\blue{0.2738}
	\\
	\noalign{\vskip 2pt}\cdashline{1-7}[.4pt/1pt]\noalign{\vskip 2pt}
	Graph WaveNet~\cite{wu_graph_2019}
	&0.1812	&0.2164	&0.2630
	&0.1492	&0.1304	&0.1706
	\\
	STGCN~\cite{yan_spatial_2018}
	&0.1925	&0.1994	&0.2380
	&0.1369	&\green{0.1549}	&\green{0.1592}
	\\
	GGRNN~\cite{ruiz_gated_2020}
	&0.1953	&0.2372	&0.2449
	&0.1403	&0.1773	&0.1836
	\\
	DCRNN~\cite{li_diffusion_2017}
	&0.2143	&0.2537 &0.2649
	&0.1702	&0.1974	&0.2104
	\\
	GTCNN~\cite{sabbaqi_gtcnn}
	&0.1843	&0.1938	&0.2411
	&\green{0.1310}	&0.1629	&0.1789
	\\
	GKNet (ours)
	&\green{0.1524}	&\green{0.1726}	&\green{0.1758}
	&0.1706	&0.1700	&0.1718
	\\[2pt]
	\hline\hline
\end{tabular}
\renewcommand{\arraystretch}{1}
\end{table*}

\begin{figure}[t]
	\centering
	\begin{tikzpicture}
    \begin{axis}[
    width=\linewidth,
    height=0.75\linewidth,
    axis line style={black},
    legend cell align={left},
    legend style={draw=black},
    tick align=outside,
    x grid style={dashed,black!60},
    xlabel={Observation ratio (\%)},
    xmajorticks=true,
    xmin=0, xmax=75,
    xtick style={color=black},
    y grid style={dashed,black!60},
    ylabel={nRMSE},
    ymajorticks=true,
    ymin=0.1, ymax=1,
    ytick style={black},
    xtick align=inside,
    ytick align=inside,
    grid = both,
    legend image post style={sharp plot,|-|}
    ]
    \addplot [mark =*,line width=1pt,pennlighterblue, mark size=1pt]
    table {%0
    5     0.899583
    10    0.898949
    15    0.801613
    20    0.801233
    25    0.746856
    30    0.708069
    35    0.687918
    40    0.599399
    45    0.458518
    50    0.380319
    55    0.309227
    60    0.233677
    65    0.171975
    };
    \addlegendentry{G-WaveNet}
    
    \addplot [mark =*,line width=1pt,pennlightergreen, mark size=1pt]
    table {%1
    5     0.843981
    10    0.809257
    15    0.826369
    20    0.794869
    25    0.649345
    30    0.634354
    35    0.578174
    40    0.491518
    45    0.411726
    50    0.358407
    55    0.233795
    60    0.181227
    65    0.162970
    };
    \addlegendentry{STGCN}
    
    \addplot [mark=*,line width=1pt,pennlighterred, mark size=1pt]
    table {%2
    5     0.873953
    10    0.776591
    15    0.742637
    20    0.703095
    25    0.690315
    30    0.602730
    35    0.581919
    40    0.552254
    45    0.457060
    50    0.478501
    55    0.396395
    60    0.249113
    65    0.186173
    };
    \addlegendentry{GGRNN}
    
    \addplot [mark=*,line width=1pt,pennblue, mark size=1pt]
    table {%3
    5     0.730809
    10    0.641221
    15    0.783412
    20    0.662764
    25    0.648229
    30    0.532714
    35    0.581634
    40    0.578871
    45    0.517965
    50    0.432721
    55    0.426531
    60    0.298985
    65    0.215513
    };
    \addlegendentry{DCRNN}
    
    \addplot [mark=*,line width=1pt,penngreen, mark size=1pt]
    table {%4
    5     0.789666
    10    0.777529
    15    0.714160
    20    0.669026
    25    0.662567
    30    0.615895
    35    0.467327
    40    0.451614
    45    0.436743
    50    0.309804
    55    0.279033
    60    0.229863
    65    0.179214
    };
    \addlegendentry{GTCNN}
    
    \addplot [mark=*,line width=1pt,pennred, mark size=1pt]
    table {%5
    5     0.285768
    10    0.269940
    15    0.293441
    20    0.261238
    25    0.281457
    30    0.254547
    35    0.242828
    40    0.215653
    45    0.196930
    50    0.196086
    55    0.187224
    60    0.178822
    65    0.176071
    };
    \addlegendentry{GKNet (ours)}
    
    \end{axis}
    
    \end{tikzpicture}
    
	\caption{The performance of data-driven models on weather temperature interpolation in a low data regime. The experiments are conducted on NOAA dataset for different amounts of missing values.}
	\label{fig:weather_low_obs}
\end{figure}
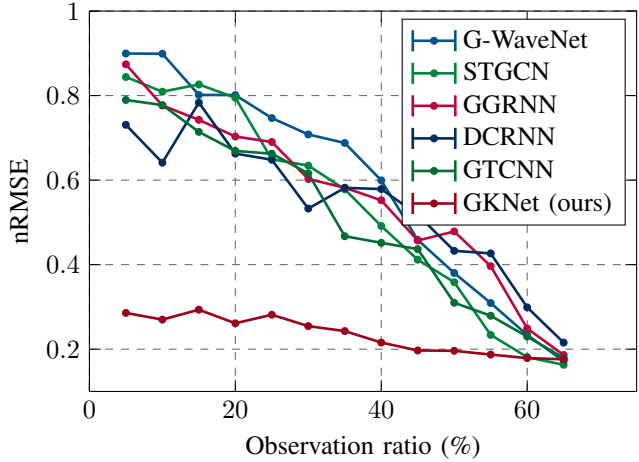

% \medskip
\noindent
\textbf{Performance in low data regime.}
5\% of the data is used for the training and validation of the models while less than 70\% of the nodes are observed.
Fig.~\ref{fig:weather_low} shows the performance of the proposed model on the NOAA dataset for $p = 90\%$ missing values over different training data ratios.
The proposed model keeps up the performance independent of the training ratio while the alternative models diverge.
In Fig.~\ref{fig:weather_low_obs}, the performance of the proposed model is evaluated for different amounts of missing values.
The results indicate that the proposed model is robust to the amount of missing values in the data, and it can handle the interpolation task even with observing a small portion of the data.
The reason behind this robustness is the ability of the model to map the data into a well-behaved domain up to an uncertainty level which is estimated during the training phase.
Hence, it allows the model to reconstruct missing values with a high precision as long as the encoder imitates the observation mapping well.

% figure for low data regime traffic forecasting
\begin{figure}[t]
	\centering
	\begin{tikzpicture}
    \begin{axis}[
    width=\linewidth,
    height=0.75\linewidth,
    axis line style={black},
    legend cell align={left},
    legend style={draw=black},
    tick align=outside,
    x grid style={dashed,black!60},
    xlabel={Training data ratio (\%)},
    xmajorticks=true,
    xmin=0, xmax=75,
    xtick style={color=black},
    y grid style={dashed,black!60},
    ylabel={nRMSE},
    ymajorticks=true,
    ymin=0.1, ymax=1,
    ytick style={black},
    xtick align=inside,
    ytick align=inside,
    grid = both,
    legend image post style={sharp plot,|-|}
    ]
    \addplot [mark =*,line width=1pt,pennlighterblue, mark size=1pt]
    table {%0
    5     0.744351
    10    0.745634
    15    0.684556
    20    0.608051
    25    0.601703
    30    0.525166
    35    0.388122
    40    0.281317
    45    0.261900
    50    0.232468
    55    0.235222
    60    0.180415
    65    0.150023
    };
    \addlegendentry{G-WaveNet}
    
    \addplot [mark =*,line width=1pt,pennlightergreen, mark size=1pt]
    table {%1
    5     0.768146
    10    0.651380
    15    0.624494
    20    0.567608
    25    0.548893
    30    0.526995
    35    0.445073
    40    0.442496
    45    0.342969
    50    0.241144
    55    0.186084
    60    0.172611
    65    0.168243
    };
    \addlegendentry{STGCN}
    
    \addplot [mark=*,line width=1pt,pennlighterred, mark size=1pt]
    table {%2
    5     0.775854
    10    0.739508
    15    0.691186
    20    0.651733
    25    0.631460
    30    0.591893
    35    0.503365
    40    0.464441
    45    0.479687
    50    0.427749
    55    0.310992
    60    0.228680
    65    0.205422
    };
    \addlegendentry{GGRNN}
    
    \addplot [mark=*,line width=1pt,pennblue, mark size=1pt]
    table {%3
    5     0.787322
    10    0.761043
    15    0.633606
    20    0.497501
    25    0.489425
    30    0.450636
    35    0.398821
    40    0.405125
    45    0.399650
    50    0.369875
    55    0.306502
    60    0.275087
    65    0.226952
    };
    \addlegendentry{DCRNN}
    
    \addplot [mark=*,line width=1pt,penngreen, mark size=1pt]
    table {%4
    5     0.778869
    10    0.728638
    15    0.728279
    20    0.668138
    25    0.514107
    30    0.491382
    35    0.474442
    40    0.361765
    45    0.344245
    50    0.321670
    55    0.264368
    60    0.210905
    65    0.185708
    };
    \addlegendentry{GTCNN}
    
    \addplot [mark=*,line width=1pt,pennred, mark size=1pt]
    table {%5
    5     0.249762
    10    0.239185
    15    0.238305
    20    0.226062
    25    0.215219
    30    0.214608
    35    0.207506
    40    0.200165
    45    0.196918
    50    0.194859
    55    0.191417
    60    0.187561
    65    0.181849
    };
    \addlegendentry{GKNet (ours)}
    
    \end{axis}
    
    \end{tikzpicture}
    
	\caption{The performance of data-driven models on weather temperature interpolation in a low data regime. The experiments are conducted on NOAA dataset for $p = 10\%$ missing values.}
	\label{fig:weather_low}
\end{figure}
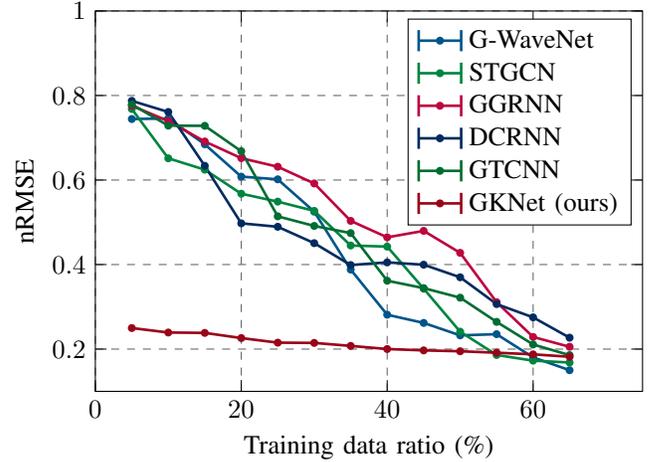

% figure for low data regime traffic forecasting

% dynamical system tracking
\subsection{Tracking  Dynamical Systems}

We now evaluate the ability of the GKNet to track the trajectory of dynamical systems without knowing the state model.
Assume a graph-based dynamical system with state variable $\bbx_t \in \reals^{N}$ and observations $\bby_{\text{obs},t} \in \reals^{N_{\text{obs}}}$ as
\begin{align}
	\bbx_{t+1} = f(\bbL, \bbx_t) + \bbw_t, \quad \bby_{\text{obs},t} = \bbM \bbH \bbx_t + \bbv_t,
\end{align}
where $f(\cdot)$ is the system dynamics, $\bbL$ is the graph Laplacian, $\bbH$ is the observation matrix, $\bbM \in \{0,1\}^{N_o\times N}$ is the masking matrix, and $\bbw_t$ and $\bbv_t$ are the process and observation noise, respectively.
The aim is to estimate the state variable $\bbx_t$ given the partial observations $\bby_{\text{obs},t}$ and track the complete the trajectory of the dynamical system $\bby_t \in \reals^N$.
This task is usually performed when the system dynamics $f(\cdot)$ are known, however, we consider the case where these dynamics need to be estimated.

% dataset
% \medskip\noindent\green{
% \framebox[\columnwidth]{\parbox{.9\linewidth}{Datasets}}}\medskip

The experiments are conducted on two synthetic datasets with the setup of~\cite{buchnik_gspkalman_2023} since it provides grounds for a more fair comparison with a method that considers known dynamics.
The graph is randomly generated with $N = 32$ nodes based on the Erdos-Renyi model.
In the first dataset, the \emph{linear} state evolution is modeled as
\begin{align}
	f(\bbL, \bbx_t) = \bbL \bbx_t.
\end{align}
For the second dataset, the \emph{nonlinear} state evolution is modeled as
\begin{align}
	f(\bbL, \bbx_t) = \sin(\bbx_t) + \cos(\bbL \bbx_t).
\end{align}
The observation matrix $\bbH$ is a graph filter with a low-pass frequency response.
The process and observation noise are Gaussian with zero mean and covariance matrices $\bbQ = \sigma_q^2\bbI_N$ and $\bbR = \sigma_r^2\bbI_{N_o}$, respectively.
The ratio of the process and observation noise is fixed $\sigma_q^2/\sigma_r^2 = 0.1$.
For both datasets, 2000 trajectories are generated with 200 time steps each.

% experimental setup

The encoder and decoder are GCNNs with three layers and second-order filters operating on the normalized Laplacian matrix of the graph.
The number of features in the middle layers are $F_1 = 16$ and $F_2 = 8$.
The objective function is the regularized mean squared error (MSE) with an uncertainty-based regularizer with
regularization weight $\lambda = 0.025$.
The evaluation metric is the normalized root mean square error (nRMSE).

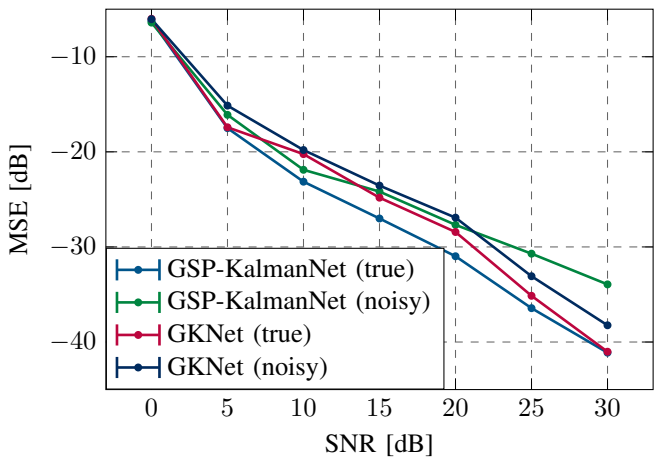
\begin{figure}[t]
	\centering
	\begin{tikzpicture}
    \begin{axis}[
    width=\linewidth,
    height=0.75\linewidth,
    axis line style={black},
    legend cell align={left},
    legend style={draw=black,at={(0,0)},anchor=south west},
    tick align=outside,
    x grid style={dashed,black!60},
    xlabel={SNR [dB]},
    xmajorticks=true,
    xmin=-3, xmax=33,
    xtick style={color=black},
    y grid style={dashed,black!60},
    ylabel={MSE [dB]},
    ymajorticks=true,
    ymin=-45, ymax=-5,
    ytick style={black},
    xtick align=inside,
    ytick align=inside,
    grid = both,
    legend image post style={sharp plot,|-|}
    ]
    \addplot [mark =*,line width=1pt,pennlighterblue, mark size=1pt]
    table {%0
    0   -6.22
    5   -17.51
    10  -23.14
    15  -27.01
    20  -30.98
    25  -36.44
    30  -41.12
    };
    \addlegendentry{GSP-KalmanNet (true)}
    
    \addplot [mark =*,line width=1pt,pennlightergreen, mark size=1pt]
    table {%1
    0   -6.41
    5   -16.11
    10  -21.89
    15  -24.18
    20  -27.68
    25  -30.71
    30  -33.94
    };
    \addlegendentry{GSP-KalmanNet (noisy)}
    
    \addplot [mark=*,line width=1pt,pennlighterred, mark size=1pt]
    table {%2
    0   -6.01
    5   -17.42
    10  -20.24
    15  -24.82
    20  -28.43
    25  -35.14
    30  -41.02
    };
    \addlegendentry{GKNet (true)}
    
    \addplot [mark=*,line width=1pt,pennblue, mark size=1pt]
    table {%3
    0   -6.05
    5   -15.14
    10  -19.81
    15  -23.55
    20  -26.92
    25  -33.08
    30  -38.24
    };
    \addlegendentry{GKNet (noisy)}
    
    \end{axis}
    
    \end{tikzpicture}
    
	\caption{The performance of data-driven models on tracking graph signals for different amounts of noise energy in a linear dynamic system.}
	\label{fig:dynamic_linear}
\end{figure}

% \medskip
\noindent
\textbf{Performance on linear dynamical system.}
We performed experiments under different amounts of system noises in two scenarios, namely, true and noisy graphs.
As Fig.~\ref{fig:dynamic_linear} indicates, the proposed model performs closely to the alternative when the true graph is known even though it does not have access to the system dynamics.
However, when the true graph is not available and a noisy estimation of it is in hand, the performance of the proposed model does not drop significantly thanks to the encoder and decoder modules that are designed to handle uncertainty in the dynamics. 

% figure for tracking linear dynamic system

\begin{figure}[t]
	\centering
	\smallskip
	\begin{tikzpicture}
    \begin{axis}[
    width=\linewidth,
    height=0.75\linewidth,
    axis line style={black},
    legend cell align={left},
    legend style={draw=black, at={(1,1)},anchor=north east},
    tick align=outside,
    x grid style={dashed,black!60},
    xlabel={SNR [dB]},
    xmajorticks=true,
    xmin=-3, xmax=33,
    xtick style={color=black},
    y grid style={dashed,black!60},
    ylabel={MSE [dB]},
    ymajorticks=true,
    ymin=-40, ymax=-12,
    ytick style={black},
    xtick align=inside,
    ytick align=inside,
    grid = both,
    legend image post style={sharp plot,|-|}
    ]
    \addplot [mark =*,line width=1pt,pennlighterblue, mark size=1pt]
    table {%0
    0   -16.24
    5   -23.41
    10  -28.94
    15  -32.00
    20  -34.19
    25  -36.42
    30  -37.81
    };
    \addlegendentry{GSP-KalmanNet (true)}
    
    \addplot [mark =*,line width=1pt,pennlightergreen, mark size=1pt]
    table {%1
    0   -15.41
    5   -22.25
    10  -27.32
    15  -31.82
    20  -33.14
    25  -34.41
    30  -34.74
    };
    \addlegendentry{GSP-KalmanNet (noisy)}
    
    \addplot [mark=*,line width=1pt,pennlighterred, mark size=1pt]
    table {%2
    0   -16.04
    5   -22.91
    10  -27.84
    15  -32.23
    20  -33.19
    25  -35.47
    30  -37.61
    };
    \addlegendentry{GKNet (true)}
    
    \addplot [mark=*,line width=1pt,pennblue, mark size=1pt]
    table {%3
    0   -16.14
    5   -22.51
    10  -28.53
    15  -32.43
    20  -32.69
    25  -35.87
    30  -36.99
    };
    \addlegendentry{GKNet (noisy)}
    
    \end{axis}
    
    \end{tikzpicture}
    
	\caption{The performance of data-driven models on tracking graph signals for different amounts of noise energy in a nonlinear dynamic system.}
	\label{fig:dynamic_nonlinear}
\end{figure}
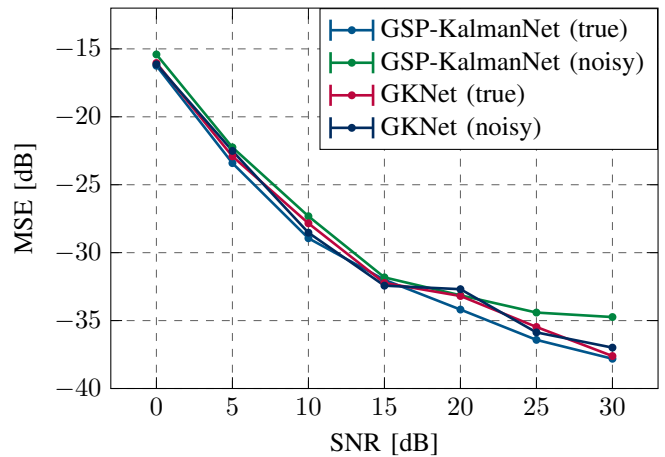

% \medskip
\noindent
\textbf{Performance on the nonlinear dynamical system.}
The experiments are conducted under different amounts of system noises for the true and noisy graph.
As shown in Fig.~\ref{fig:dynamic_nonlinear}, the proposed model performs competitively with the alternative model when the true graph is known while it is blind to the system dynamics.
Again, the proposed model shines when a noisy estimation of the graph is replaced by inferring an uncertain evolution process for the state variable $\bbx_t$ through the encoder and decoder modules.
The results illustrate that the performance gap for the GKNet model is smaller in the nonlinear scenario compared to the linear one.
We attribute this to the fact that the nonlinear dynamics violate Kalman recursions validity, and the GKNet is flexible enough to capture these dynamics by learning a nonlinear mapping from the observations to the state variable.

% figure for tracking nonlinear dynamic system

\section{Case Study in Water Networks}
\label{sec_casestudy}
% -----------------------------------------------------------------------------------------------
% CASE STUDY: WATER NETWORKS 
% -----------------------------------------------------------------------------------------------

% -----------------------------------------------------------------------------------------------
%  motivation

We now show an interesting application of the proposed approach in a stormwater system (SWS) network.
Stormwater networks are water networks designed to collect and transport stormwater runoff to prevent flooding in urban areas.
Monitoring and predicting the hydraulic head (pressure) in the drainage system is essential for for estimating the state of the systems and improving its performance through forecasting and real-time control~\cite{mullapudi_deep_2020}.
The hydraulic head at nodes in a stormwater system typically equals the water surface elevation, except when surcharging occurs and pressure builds up.
While physical-based methods are conventional to solve such task, data-driven solutions are increasingly adopted for computational efficiency and to avoid difficult parameter estimation of numerical models~\cite{garzon_transferable_2024,fu_making_2024}.
Yet, using data-driven models for stormwater networks comes with challenges.
First, the number of sensors in the system is limited due to the cost of the monitoring infrastructure and its maintenance.
Second, viable data-driven models require lots of data, and this is not always available, especially if the system has been recently installed.
Third, the hydraulic head (and all the variables of interest in the system) are affected by the stormwater network topology that needs to be incorporated in the model.
Hence, model-based approaches are handy in such task as they can incorporate the network topology and the physical laws governing the system.
In this case study, we demonstrate the effectiveness of the proposed GKNet model in predicting the hydraulic head in a water network using a limited number of sensors and training samples.

% -----------------------------------------------------------------------------------------------
%  problem formulation

% \medskip
\noindent\textbf{Problem formulation: }
The graph $\ccalG$ represents the physical topology of the water network, where each node corresponds to a junction (like manholes or storage nodes), each edge is a pipe, and the edge weights are a measure of the pipe conductance.
Let $N$ be the number of the nodes and $M$ be the number of edges in the graph.
The time-varying graph signal $\bby_t \in \reals^N$ is the hydraulic head at each junction at time $t$, however, we can only observe this value on a subset of the nodes due to sensor limited availability.
Hence, we define $\tby_t = \bbM \bby_t \in \reals^{N_o}$ as the observed hydraulic head at time $t$, where $\bbM \in \{0,1\}^{N_o \times N}$ is the observation matrix and $N_o$ is the number of observed junctions in the network.
Another observed value in such system is the rainfall at each node, which can be indirectly represented in runoff signal (i.e., water flows entering the sewer system) denoted by $\bbr_t \in \reals^N$ available at all nodes in the network.
The latter can be seen as an additional input graph signal entering the system and getting coupled with the hydraulic head.
The goal is to predict the hydraulic head at all nodes in the network $\bby_{t+1}$ given the observed hydraulic head $\tby_t$, the runoff signal $\bbr_t$, and the graph structure $\ccalG$.

% -----------------------------------------------------------------------------------------------
%  state space model

% \medskip
\noindent\textbf{Modeling the state space: }
We model the state space of the hydraulic head similar to~\eqref{eq:state_space_model}, however, the runoff signal $\bbr_t$ should be considered as an input to the state space model.
Thus, the state equation becomes
\begin{equation}
    \bbx_{t+1} = -c\bbL\bbx_{t} + \bbH_{in}(\bbL)\bbr_t + \bbB \diag(\bbalpha)\bbw_t
    \label{eq:state_eq_water}
\end{equation}
where $\bbH_{in}(\bbL)$ is a graph filter incorporating the effect of the input to the diffusion of our hidden latent variable $\bbx_t$.
The observation equation is defined as
\begin{equation}
    \tby_t = \bbM \bbH(\bbL)\bbx_t + \bbv_t
\end{equation}
where $\bbM$ is the binary observation matrix and known due to the sensor placement, and $\bby_t = \bbH(\bbL)\bbx_t$ can be used for the prediction and imputation of the data over the whole network.

% -----------------------------------------------------------------------------------------------
%  modifications

% \medskip
\noindent\textbf{GKNet modifications: }
The GKNet model can be easily modified to incorporate the runoff signal as an input to the state space model.
The input graph filter $\bbH_{in}(\bbL)$ can be learned from the data similar to the Kalman gain $\hbK_t$ and edge uncertainties $\bbalpha$ in the inference module.
The Kalman module needs to be modified in the prediction step to incorporate the runoff signal as
\begin{equation}
    \bbx^-_{t+1} = -c\bbL\bbx_t + \bbH_{in}(\bbL)\bbr_t
\end{equation}
where the diffusivity $c$ and input graph filter $\bbH_{in}(\bbL)$ are learned from the data the inference module.
The Kalman update does not change as the deterministic runoff input does not affect the model statistics.
Moreover, the regularization term in the loss function~\eqref{eq:loss} should be modified to include the runoff signal effect in the smoothness of the state variable $\bbx_t$ as
\begin{equation}
    \ccalL_t = \|\hby_t-\bby_t\|_2^2 + \lambda \bbepsilon_t^T\bbQ^{-1}\bbepsilon_t
\end{equation}
with $\bbepsilon_t = \bbx_{t+1} + c\bbL\bbx_t - \bbH_{in}(\bbL)\bbr_t$ as the deviation of the state variable $\bbx_t$ from the diffusion process~\eqref{eq:state_eq_water}.

% -----------------------------------------------------------------------------------------------
%  dataset

% \medskip
\noindent\textbf{Dataset: }
We use the dataset in~\cite{garzon_transferable_2024} to design the experiments.
The dataset is derived from a calibrated numerical model for the urban drainage system of Tuindorp, in the city of Utrecht, Netherlands.
The dataset contains the hydraulic head and runoff signals at each junction in the water network for 160 (125 real, 35 synthetic) rainfall events.
The length of each event varies between 6 and 72 hours (24 hours long on average) with a sampling rate of 5 minutes.

% -----------------------------------------------------------------------------------------------
%  exp1: number of sensors

% figure for the performance of on the water network based on observation ratio
\begin{figure}[t]
	\centering
	\begin{tikzpicture}
    \begin{axis}[
    width=\linewidth,
    height=0.75\linewidth,
    axis line style={black},
    legend cell align={left},
    legend style={draw=black},
    tick align=outside,
    x grid style={dashed,black!60},
    xlabel={Observation ratio (\%)},
    xmajorticks=true,
    xmin=0, xmax=105,
    xtick style={color=black},
    y grid style={dashed,black!60},
    ylabel={nRMSE},
    ymajorticks=true,
    ymin=0.01, ymax=1,
    ytick style={black},
    xtick align=inside,
    ytick align=inside,
    grid = both,
    legend image post style={sharp plot,|-|}
    ]
    \addplot [mark =*,line width=1pt,pennlighterblue, mark size=1pt]
    table {%0
    5      0.998351
    10     0.988978
    15     0.931832
    20     0.900781
    25     0.888454
    30     0.845336
    35     0.816460
    40     0.718487
    45     0.659602
    50     0.636378
    55     0.622875
    60     0.610112
    65     0.514570
    70     0.482484
    75     0.433038
    80     0.329454
    85     0.245266
    90     0.242688
    95     0.222044
    100    0.141112
    };
    \addlegendentry{MLP}

    \addplot [mark=*,line width=1pt,pennlighterred, mark size=1pt]
    table {%1
    5     0.870339
    10    0.716413
    15    0.699391
    20    0.696903
    25    0.684270
    30    0.614332
    35    0.597013
    40    0.583546
    45    0.580946
    50    0.560585
    55    0.544704
    60    0.538558
    65    0.488577
    70    0.477542
    75    0.343323
    80    0.200208
    85    0.166885
    90    0.125758
    95    0.113793
    100   0.067558 
    };
    \addlegendentry{GNN}

    \addplot [mark =*,line width=1pt,pennlightergreen, mark size=1pt]
    table {%2
    5      0.321530
    10     0.320304
    15     0.314807
    20     0.325266
    25     0.303864
    30     0.289020
    35     0.269430
    40     0.227537
    45     0.278312
    50     0.226936
    55     0.225101
    60     0.215176
    65     0.266277
    70     0.239854
    75     0.285345
    80     0.254244
    85     0.213277
    90     0.196824
    95     0.190199
    100    0.127910
    };
    \addlegendentry{GKNet}
    
    \end{axis}
    
    \end{tikzpicture}
    
	\caption{The performance of the GKNet model in predicting the hydraulic head in the water network using a limited number of sensors compared to the baselines.}
	\label{fig:water_net_obs}
\end{figure}
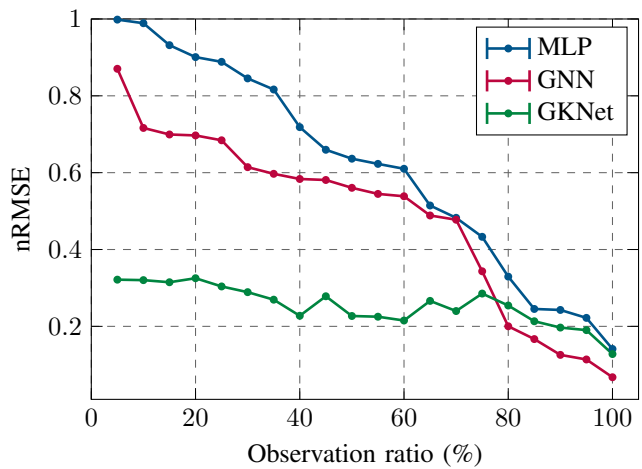

% \medskip
\noindent\textbf{Results- number of sensors: }
We evaluate the GKNet model in predicting the hydraulic head at all junctions in the water network using a limited number of sensors.
We compare with two standard baselines in the domain: a multi-layer perceptron (MLP) and a graph neural network (GNN) similar to~\cite{garzon_transferable_2024}.
The experiment is conducted on different amount of observation ratios ranging from 5\% to 100\% of the junctions.
The hyperparameters of each experiment are grid searched separately for the minimum nRMSE on the validation set.
As the MLP and GNN baselines are not designed for data with missing values, a Tikhonov regularized smoothed version of the data is passed through them as the input.
The results in Fig.~\ref{fig:water_net_obs} show that the GKNet model outperforms the baselines in predicting the hydraulic head in the water network with limited number of sensors, while the performance is still competitive when all nodes are observed.
This is due to the GKNet model's ability to incorporate the network topology to derive a latent space where the data can be approximated by few parameters.

% -----------------------------------------------------------------------------------------------
%  exp2: data efficiency

% figure for the performance of on the water network based on rainfall events
\begin{figure}[t]
	\centering
	\begin{tikzpicture}
    \begin{axis}[
    width=\linewidth,
    height=0.75\linewidth,
    axis line style={black},
    legend cell align={left},
    legend style={draw=black},
    tick align=outside,
    x grid style={dashed,black!60},
    xlabel={Number of rainfall events},
    xmajorticks=true,
    xmin=0, xmax=105,
    xtick style={color=black},
    y grid style={dashed,black!60},
    ylabel={nRMSE},
    ymajorticks=true,
    ymin=0.01, ymax=1,
    ytick style={black},
    xtick align=inside,
    ytick align=inside,
    grid = both,
    legend image post style={sharp plot,|-|}
    ]
    \addplot [mark =*,line width=1pt,pennlighterblue, mark size=1pt]
    table {%0
    5      0.790725
    10     0.788737
    15     0.763333
    20     0.758692
    25     0.699546
    30     0.652890
    35     0.646656
    40     0.636477
    45     0.512644
    50     0.506524
    55     0.475672
    60     0.394866
    65     0.409176
    70     0.420010
    75     0.361170
    80     0.296393
    85     0.294511
    90     0.245293
    95     0.161905
    100    0.102542
    };
    \addlegendentry{MLP}

    \addplot [mark=*,line width=1pt,pennlighterred, mark size=1pt]
    table {%1
    5      0.362848
    10     0.314843
    15     0.325454
    20     0.280455
    25     0.304887
    30     0.303968
    35     0.284076
    40     0.262392
    45     0.309336
    50     0.280879
    55     0.275041
    60     0.222301
    65     0.218203
    70     0.201823
    75     0.194010
    80     0.163052
    85     0.190126
    90     0.142057
    95     0.135628
    100    0.108564
    };
    \addlegendentry{GNN}

    \addplot [mark =*,line width=1pt,pennlightergreen, mark size=1pt]
    table {%2
    5      0.378448
    10     0.343794
    15     0.314228
    20     0.297114
    25     0.281299
    30     0.266595
    35     0.265137
    40     0.260971
    45     0.270563
    50     0.270834
    55     0.247738
    60     0.231557
    65     0.255701
    70     0.230604
    75     0.212169
    80     0.216224
    85     0.182806
    90     0.196115
    95     0.187483
    100    0.136112
    };
    \addlegendentry{GKNet}
    
    \end{axis}
    
    \end{tikzpicture}
    
	\caption{The performance of the GKNet model in predicting the hydraulic head in the water network using a limited number of rainfall events compared to the baselines.}
	\label{fig:water_net_event}
\end{figure}
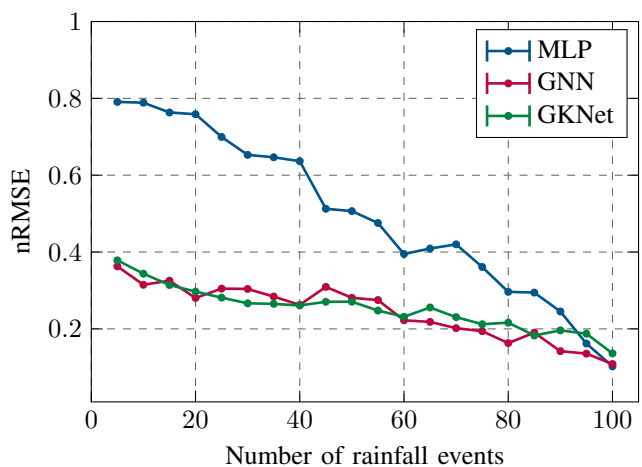

% \medskip
\noindent\textbf{Results- data efficiency: }
In this experiment, we evaluate the model based on the available number of rain events in the training set varying from 5 to 100 given the measurments over 80\% of the junctions.
The lack of rainfall events may, for instance, reflect a scenario where the monitoring system was just installed, and we aim to deploy the model as soon as possible rather than waiting to accumulate a larger dataset
Fig.~\ref{fig:water_net_event} shows that the GNN model is already efficient in this scenario while the MLP fails to exploit any patterns in the data when a limited number of samples is in hand.
The GKNet model performs best along with GNN in the extreme low data regime becasue they both exploit the patterns in the data.
This is due to the fact that the temporal behavior of the data is strongly governed by the topology of the network; hence, when the hydraulic head is known for most nodes, the GNN itself can capture the temporal patterns in the data.

% -----------------------------------------------------------------------------------------------
%  exp2: transferability

\begin{table*}[t]
	\centering
	\caption{The transferability of GKNet compared to GNN for different ratios of observed nodes in the water network.}
	\label{table:water_net_1}
	\newcolumntype{C}{>{\centering\let\newline\\\arraybackslash\hspace{0pt}}m{1cm}}

\renewcommand{\arraystretch}{1.25}
\begin{tabular}{m{2.75cm}CCC|CCC}
	\hline\hline
	\multirow{2}{1cm}{\centering nRMSE}
	&\multicolumn{3}{c}{in-domain}
	&\multicolumn{3}{c}{transfer domain}
	\\ \cline{2-7}
	&$100\%$	&$50\%$	&$20\%$
	&$100\%$	&$50\%$	&$20\%$
	\\ \hline
	GNN~\cite{garzon_transferable_2024}
	&\green{0.0675}	&0.5605	&0.6970	
	&0.1344	&0.6172	&0.7542
	\\
	GKNet~($\bbH_z = \bbzero$)
	&0.1408	&0.2441	&0.3396	
	&0.1899	&0.2562	&0.3719
	\\
	GKNet~(fine-tune $\bbH_z$)
	&0.1279	&\green{0.2269}	&\green{0.3252}	
	&\green{0.1291}	&\green{0.2305}	&\green{0.3420}
	\\[2pt]
	\hline\hline
\end{tabular}
\renewcommand{\arraystretch}{1}
\end{table*}

% \medskip
\noindent\textbf{Results- transferability: }
We now use the trained neural network over the measured domain to predict the hydraulic head in the unseen/transfer domain. In particular, we follow the approach in~\cite{garzon_transferable_2024}, where models are trained on one part of the Tuindorp network and transferred to another part.
The ability of a data-driven model to transfer across sensor settings and domains is particularly important in water networks due to sensor reconfigurations or data scarcity, as we aim to reuse knowledge acquired in other contexts~\cite{kerimov_towards_2024}. 
Note that as the $\bbH_z$ in~\eqref{eq:gru} depends on the number of graph nodes, the GKNet model is not generally transferable to a domain with a different number of nodes.
However, one can remove the dependency of the recurrent unit to $\bbsigma_t$ by inserting $\bbH_z = \bbzero$ or re-train this matrix on the new domain.
Both cases are investigated in this experiment.
The results in Table~\ref{table:water_net_1} suggests that the GKNet model benefits from similar transferability properties as the GNN model even in scenarios with a limited number of sensors in training samples.
However, fine-tuning the $\bbH_z$ matrix on the new domain improves the transferability of the GKNet model significantly as it adapts the recurrent transformation of the model to the data behavior in the new domain.
In conclusion, as the main improvement over GNNs, both models propose a satisfactory performance in the transfer domain even when the observed nodes are limited in the network which is a common scenario in real-world applications. 

% -----------------------------------------------------------------------------------------------
% END: CASE STUDY: WATER NETWORKS 
% -----------------------------------------------------------------------------------------------

%%%---------------------------------------------------------------------

%%%---------------------------------------------------------------------
\section{Conclusion}
\label{sec_conc}
This paper presented a novel method for inferring time varying data over graphs by unrolling the graph Kalman filter algorithm in a state space identification scheme.
The goal is to transform the observed data into a latent space where the state variable evolves according to a diffusion process over the graph.
We proposed a state equation based on stochastic partial differential equations which follows a heat diffusion process over the graph up to an uncertainty level locally around graph edges.
This flexible state representation allows the observation model to map the observed data into this well-behaved latent space with more degrees of freedom.
Initially, we solved this inference problem via the EM algorithm where the E-step is a Kalman filter and the M-step is solvable with gradient-based optimization techniques.
Unlike deep learning models, this approach works well even with few data sample, yet, it suffers from scalability issues, linear expressivity, and strict stochastic assumptions.
To address these issues, we proposed a model-based deep learning approach which uses a graph-based encoder and decoder to transform data into the latent space and vice versa.
In the latent space, we unroll the Kalman filter and learn its parameters through a recurrent recognition model.
The proposed model outperforms both statistical and deep learning models in different tasks when a limited amount of data is available while maintaining its performance competitive to deep learning models when the data is abundant.
Moreover, we demonstrated the effectiveness of the proposed model in a case study on water networks imputing the hydraulic head time series in a water network using a limited number of sensors and training samples.
%%%---------------------------------------------------------------------

%%%---------------------------------------------------------------------
% references section
% \newpage
\bibliographystyle{IEEEtran}
\bibliography{mystyle/myIEEEabrv,unrolling}

% Generated by IEEEtran.bst, version: 1.14 (2015/08/26)
\begin{thebibliography}{10}
\providecommand{\url}[1]{#1}
\csname url@samestyle\endcsname
\providecommand{\newblock}{\relax}
\providecommand{\bibinfo}[2]{#2}
\providecommand{\BIBentrySTDinterwordspacing}{\spaceskip=0pt\relax}
\providecommand{\BIBentryALTinterwordstretchfactor}{4}
\providecommand{\BIBentryALTinterwordspacing}{\spaceskip=\fontdimen2\font plus
\BIBentryALTinterwordstretchfactor\fontdimen3\font minus \fontdimen4\font\relax}
\providecommand{\BIBforeignlanguage}[2]{{%
\expandafter\ifx\csname l@#1\endcsname\relax
\typeout{** WARNING: IEEEtran.bst: No hyphenation pattern has been}%
\typeout{** loaded for the language `#1'. Using the pattern for}%
\typeout{** the default language instead.}%
\else
\language=\csname l@#1\endcsname
\fi
#2}}
\providecommand{\BIBdecl}{\relax}
\BIBdecl

\bibitem{sabbaqi_inferring}
M.~Sabbaqi and E.~Isufi, ``Inferring time varying signals over uncertain graphs,'' in \emph{ICASSP 2024 - 2024 IEEE International Conference on Acoustics, Speech and Signal Processing (ICASSP)}, 2024, pp. 9876--9880.

\bibitem{kolaczyk_statistical_2014}
E.~D. Kolaczyk, G.~Cs{\'a}rdi, E.~D. Kolaczyk, and G.~Cs{\'a}rdi, ``Statistical models for network graphs,'' \emph{Statistical analysis of network data with R}, pp. 85--109, 2014.

\bibitem{ortega_overview_2018}
A.~Ortega, P.~Frossard, J.~Kovačević, J.~M.~F. Moura, and P.~Vandergheynst, ``Graph signal processing: Overview, challenges, and applications,'' \emph{Proceedings of the IEEE}, vol. 106, no.~5, pp. 808--828, 2018.

\bibitem{marisca_learning_2022}
I.~Marisca, A.~Cini, and C.~Alippi, ``Learning to reconstruct missing data from spatiotemporal graphs with sparse observations,'' in \emph{Advances in Neural Information Processing Systems}, vol.~35.\hskip 1em plus 0.5em minus 0.4em\relax Curran Associates, Inc., 2022, pp. 32\,069--32\,082.

\bibitem{yi_fouriergnn_nodate}
K.~Yi, Q.~Zhang, W.~Fan, H.~He, L.~Hu, P.~Wang, N.~An, L.~Cao, and Z.~Niu, ``Fouriergnn: Rethinking multivariate time series forecasting from a pure graph perspective,'' in \emph{Advances in Neural Information Processing Systems}, vol.~36.\hskip 1em plus 0.5em minus 0.4em\relax Curran Associates, Inc., 2023, pp. 69\,638--69\,660.

\bibitem{li_deep_2023}
G.~Li and J.~J. Jung, ``Deep learning for anomaly detection in multivariate time series: Approaches, applications, and challenges,'' \emph{Information Fusion}, vol.~91, pp. 93--102, 2023.

\bibitem{giannakis_topology_2018}
G.~B. Giannakis, Y.~Shen, and G.~V. Karanikolas, ``Topology identification and learning over graphs: Accounting for nonlinearities and dynamics,'' \emph{Proceedings of the IEEE}, vol. 106, no.~5, pp. 787--807, 2018.

\bibitem{dong_learning_2019}
X.~Dong, D.~Thanou, M.~Rabbat, and P.~Frossard, ``Learning graphs from data: A signal representation perspective,'' \emph{IEEE Signal Processing Magazine}, vol.~36, no.~3, pp. 44--63, 2019.

\bibitem{mateos_connecting_2019}
G.~Mateos, S.~Segarra, A.~G. Marques, and A.~Ribeiro, ``Connecting the dots: Identifying network structure via graph signal processing,'' \emph{IEEE Signal Processing Magazine}, vol.~36, no.~3, pp. 16--43, 2019.

\bibitem{buciulea_learning_2022}
A.~Buciulea, S.~Rey, and A.~G. Marques, ``Learning graphs from smooth and graph-stationary signals with hidden variables,'' \emph{IEEE Transactions on Signal and Information Processing over Networks}, vol.~8, pp. 273--287, 2022.

\bibitem{shen_kernel_2017}
Y.~Shen, B.~Baingana, and G.~B. Giannakis, ``Kernel-based structural equation models for topology identification of directed networks,'' \emph{IEEE Transactions on Signal Processing}, vol.~65, no.~10, pp. 2503--2516, 2017.

\bibitem{isufi_graphfilters_2024}
E.~Isufi, F.~Gama, D.~I. Shuman, and S.~Segarra, ``Graph filters for signal processing and machine learning on graphs,'' \emph{IEEE Transactions on Signal Processing}, pp. 1--32, 2024.

\bibitem{ceci_graph_2020}
E.~Ceci and S.~Barbarossa, ``Graph signal processing in the presence of topology uncertainties,'' \emph{IEEE Transactions on signal processing}, vol.~68, pp. 1558--1573, 2020.

\bibitem{sabbaqi_gtcnn}
M.~Sabbaqi and E.~Isufi, ``Graph-time convolutional neural networks: Architecture and theoretical analysis,'' \emph{IEEE Transactions on Pattern Analysis and Machine Intelligence}, 2023.

\bibitem{gao_learning_2023}
Z.~Gao and E.~Isufi, ``Learning stochastic graph neural networks with constrained variance,'' \emph{IEEE Transactions on Signal Processing}, vol.~71, pp. 358--371, 2023.

\bibitem{gao_stochastic_2021}
Z.~Gao, E.~Isufi, and A.~Ribeiro, ``Stochastic graph neural networks,'' \emph{IEEE Transactions on Signal Processing}, vol.~69, pp. 4428--4443, 2021.

\bibitem{gao_stability_2021}
------, ``Stability of graph convolutional neural networks to stochastic perturbations,'' \emph{Signal Processing}, vol. 188, p. 108216, 2021.

\bibitem{chamberlain_grand_2021}
B.~Chamberlain, J.~Rowbottom, M.~I. Gorinova, M.~Bronstein, S.~Webb, and E.~Rossi, ``Grand: Graph neural diffusion,'' in \emph{International Conference on Machine Learning}.\hskip 1em plus 0.5em minus 0.4em\relax PMLR, 2021, pp. 1407--1418.

\bibitem{song_robustness_2022}
Y.~Song, Q.~Kang, S.~Wang, K.~Zhao, and W.~P. Tay, ``On the robustness of graph neural diffusion to topology perturbations,'' \emph{Advances in Neural Information Processing Systems}, vol.~35, pp. 6384--6396, 2022.

\bibitem{smola_kernels_2003}
A.~J. Smola and R.~Kondor, ``Kernels and regularization on graphs,'' in \emph{Learning Theory and Kernel Machines: 16th Annual Conference on Learning Theory and 7th Kernel Workshop, COLT/Kernel 2003, Washington, DC, USA, August 24-27, 2003. Proceedings}.\hskip 1em plus 0.5em minus 0.4em\relax Springer, 2003, pp. 144--158.

\bibitem{romero_kernelbased_2017}
D.~Romero, V.~N. Ioannidis, and G.~B. Giannakis, ``Kernel-based reconstruction of space-time functions on dynamic graphs,'' \emph{IEEE Journal of Selected Topics in Signal Processing}, vol.~11, no.~6, pp. 856--869, 2017.

\bibitem{lu_probabilistic_2021}
Q.~Lu and G.~B. Giannakis, ``Probabilistic reconstruction of spatio-temporal processes over multi-relational graphs,'' \emph{IEEE Transactions on Signal and Information Processing over Networks}, vol.~7, pp. 166--176, 2021.

\bibitem{lu_spatiotemporal_2021}
------, ``Spatio-temporal inference of dynamical gaussian processes over graphs,'' in \emph{2021 55th Asilomar Conference on Signals, Systems, and Computers}.\hskip 1em plus 0.5em minus 0.4em\relax IEEE, 2021, pp. 1515--1519.

\bibitem{zhi_gaussian_2023}
Y.-C. Zhi, Y.~C. Ng, and X.~Dong, ``Gaussian processes on graphs via spectral kernel learning,'' \emph{IEEE Transactions on Signal and Information Processing over Networks}, 2023.

\bibitem{nikitin_non_2022}
A.~V. Nikitin, S.~John, A.~Solin, and S.~Kaski, ``Non-separable spatio-temporal graph kernels via spdes,'' in \emph{International Conference on Artificial Intelligence and Statistics}.\hskip 1em plus 0.5em minus 0.4em\relax PMLR, 2022, pp. 10\,640--10\,660.

\bibitem{jin_survey_2024}
M.~Jin, H.~Y. Koh, Q.~Wen, D.~Zambon, C.~Alippi, G.~I. Webb, I.~King, and S.~Pan, ``A survey on graph neural networks for time series: Forecasting, classification, imputation, and anomaly detection,'' \emph{IEEE Transactions on Pattern Analysis and Machine Intelligence}, vol.~46, no.~12, pp. 10\,466--10\,485, 2024.

\bibitem{ruiz_gated_2020}
L.~Ruiz, F.~Gama, and A.~Ribeiro, ``Gated graph recurrent neural networks,'' \emph{IEEE Transactions on Signal Processing}, vol.~68, pp. 6303--6318, 2020.

\bibitem{shlezinger_model-based_2022}
N.~Shlezinger, Y.~C. Eldar, and S.~P. Boyd, ``Model-{Based} {Deep} {Learning}: {On} the {Intersection} of {Deep} {Learning} and {Optimization},'' \emph{IEEE Access}, vol.~10, pp. 115\,384--115\,398, 2022.

\bibitem{shlezinger_model-based_2023}
N.~Shlezinger, J.~Whang, Y.~C. Eldar, and A.~G. Dimakis, ``Model-{Based} {Deep} {Learning},'' \emph{Proceedings of the IEEE}, vol. 111, no.~5, pp. 465--499, 2023.

\bibitem{hadou_robust_2023}
S.~Hadou, N.~NaderiAlizadeh, and A.~Ribeiro, ``Robust stochastically-descending unrolled networks,'' 2023.

\bibitem{sabbaqi_gtfunrolling}
M.~Sabbaqi and E.~Isufi, ``Graph-time trend filtering and unrolling network,'' in \emph{2023 31st European Signal Processing Conference (EUSIPCO)}, 2023, pp. 1230--1234.

\bibitem{chen_graph_2021}
S.~Chen, Y.~C. Eldar, and L.~Zhao, ``Graph unrolling networks: Interpretable neural networks for graph signal denoising,'' \emph{IEEE Transactions on Signal Processing}, vol.~69, pp. 3699--3713, 2021.

\bibitem{nagahama_graph_2022}
M.~Nagahama, K.~Yamada, Y.~Tanaka, S.~H. Chan, and Y.~C. Eldar, ``Graph signal restoration using nested deep algorithm unrolling,'' \emph{IEEE Transactions on Signal Processing}, vol.~70, pp. 3296--3311, 2022.

\bibitem{chen_time_2021}
S.~Chen and Y.~C. Eldar, ``Time-varying graph signal inpainting via unrolling networks,'' in \emph{ICASSP 2021 - 2021 IEEE International Conference on Acoustics, Speech and Signal Processing (ICASSP)}, 2021, pp. 8092--8097.

\bibitem{vu_Unrolling_2021}
H.~Vu, G.~Cheung, and Y.~C. Eldar, ``Unrolling of deep graph total variation for image denoising,'' in \emph{ICASSP 2021 - 2021 IEEE International Conference on Acoustics, Speech and Signal Processing (ICASSP)}, 2021, pp. 2050--2054.

\bibitem{isufi_observing_2020}
E.~Isufi, P.~Banelli, P.~D. Lorenzo, and G.~Leus, ``Observing and tracking bandlimited graph processes from sampled measurements,'' \emph{Signal Processing}, vol. 177, p. 107749, 2020.

\bibitem{revach_kalman_2022}
G.~Revach, N.~Shlezinger, X.~Ni, A.~L. Escoriza, R.~J.~G. van Sloun, and Y.~C. Eldar, ``Kalmannet: Neural network aided kalman filtering for partially known dynamics,'' \emph{IEEE Transactions on Signal Processing}, vol.~70, pp. 1532--1547, 2022.

\bibitem{sagi_extended_2023}
G.~Sagi, N.~Shlezinger, and T.~Routtenberg, ``Extended {Kalman} {Filter} for {Graph} {Signals} in {Nonlinear} {Dynamic} {Systems},'' in \emph{{ICASSP} 2023 - 2023 {IEEE} {International} {Conference} on {Acoustics}, {Speech} and {Signal} {Processing} ({ICASSP})}, 2023, pp. 1--5.

\bibitem{buchnik_gspkalman_2023}
I.~Buchnik, G.~Sagi, N.~Leinwand, Y.~Loya, N.~Shlezinger, and T.~Routtenberg, ``Gsp-kalmannet: Tracking graph signals via neural-aided kalman filtering,'' \emph{arXiv preprint arXiv:2311.16602}, 2023.

\bibitem{shi_kalman_2009}
L.~Shi, ``Kalman {Filtering} {Over} {Graphs}: {Theory} and {Applications},'' \emph{IEEE Transactions on Automatic Control}, vol.~54, no.~9, pp. 2230--2234, 2009.

\bibitem{li_unscented_2023}
W.~Li, X.~Fu, B.~Zhang, and Y.~Liu, ``Unscented {Kalman} filter of graph signals,'' \emph{Automatica}, vol. 148, p. 110796, 2023.

\bibitem{su_graph-frequency_2024}
L.~Su, Z.~Han, J.~Zhao, and W.~Wang, ``Graph-frequency domain kalman filtering for industrial pipe networks subject to measurement outliers,'' \emph{IEEE Transactions on Industrial Informatics}, vol.~20, no.~5, pp. 7977--7985, 2024.

\bibitem{gama_stability_2020}
F.~Gama, J.~Bruna, and A.~Ribeiro, ``Stability properties of graph neural networks,'' \emph{IEEE Transactions on Signal Processing}, vol.~68, pp. 5680--5695, 2020.

\bibitem{levie_transferability_2019}
R.~Levie, E.~Isufi, and G.~Kutyniok, ``On the transferability of spectral graph filters,'' in \emph{2019 13th International conference on Sampling Theory and Applications (SampTA)}.\hskip 1em plus 0.5em minus 0.4em\relax IEEE, 2019, pp. 1--5.

\bibitem{cappe_hmm_2005}
O.~Cappé, E.~Moulines, and T.~Rydén, \emph{Inference in Hidden Markov Models}, ser. Springer Series in Statistics.\hskip 1em plus 0.5em minus 0.4em\relax Springer New York, NY, 2005.

\bibitem{dempster_maximum_1977}
A.~P. Dempster, N.~M. Laird, and D.~B. Rubin, ``Maximum likelihood from incomplete data via the em algorithm,'' \emph{Journal of the royal statistical society: series B (methodological)}, vol.~39, no.~1, pp. 1--22, 1977.

\bibitem{baum_statistical_1966}
L.~E. Baum and T.~Petrie, ``Statistical inference for probabilistic functions of finite state markov chains,'' \emph{The annals of mathematical statistics}, vol.~37, no.~6, pp. 1554--1563, 1966.

\bibitem{borovych_data-driven_2022}
A.~Borovykh, D.~Kalise, A.~Laignelet, and P.~Parpas, ``Data-driven initialization of deep learning solvers for hamilton-jacobi-bellman pdes,'' \emph{IFAC-PapersOnLine}, vol.~55, no.~30, pp. 168--173, 2022, 25th International Symposium on Mathematical Theory of Networks and Systems MTNS 2022.

\bibitem{kalman_new_1961}
R.~E. Kalman and R.~S. Bucy, ``New {Results} in {Linear} {Filtering} and {Prediction} {Theory},'' \emph{Journal of Basic Engineering}, vol.~83, no.~1, pp. 95--108, 1961.

\bibitem{gama_convolution_2020}
F.~Gama, E.~Isufi, G.~Leus, and A.~Ribeiro, ``Graphs, convolutions, and neural networks: From graph filters to graph neural networks,'' \emph{IEEE Signal Processing Magazine}, vol.~37, no.~6, pp. 128--138, 2020.

\bibitem{bapat_graphs_2010}
R.~B. Bapat, \emph{Graphs and matrices}.\hskip 1em plus 0.5em minus 0.4em\relax Springer, 2010, vol.~27.

\bibitem{li_diffusion_2017}
Y.~Li, R.~Yu, C.~Shahabi, and Y.~Liu, ``Diffusion convolutional recurrent neural network: Data-driven traffic forecasting,'' \emph{arXiv preprint arXiv:1707.01926}, 2017.

\bibitem{box_time_2015}
G.~E. Box, G.~M. Jenkins, G.~C. Reinsel, and G.~M. Ljung, \emph{Time series analysis: forecasting and control}.\hskip 1em plus 0.5em minus 0.4em\relax John Wiley \& Sons, 2015.

\bibitem{isufi_forecasting_2019}
E.~Isufi, A.~Loukas, N.~Perraudin, and G.~Leus, ``Forecasting time series with varma recursions on graphs,'' \emph{IEEE Transactions on Signal Processing}, vol.~67, no.~18, pp. 4870--4885, 2019.

\bibitem{wu_graph_2019}
Z.~Wu, S.~Pan, G.~Long, J.~Jiang, and C.~Zhang, ``Graph wavenet for deep spatial-temporal graph modeling,'' \emph{arXiv preprint arXiv:1906.00121}, 2019.

\bibitem{yan_spatial_2018}
S.~Yan, Y.~Xiong, and D.~Lin, ``Spatial temporal graph convolutional networks for skeleton-based action recognition,'' \emph{Proceedings of the AAAI Conference on Artificial Intelligence}, vol.~32, no.~1, Apr. 2018.

\bibitem{mullapudi_deep_2020}
A.~Mullapudi, M.~J. Lewis, C.~L. Gruden, and B.~Kerkez, ``Deep reinforcement learning for the real time control of stormwater systems,'' \emph{Advances in Water Resources}, vol. 140, p. 103600, 2020.

\bibitem{garzon_transferable_2024}
A.~Garzón, Z.~Kapelan, J.~Langeveld, and R.~Taormina, ``Transferable and data efficient metamodeling of storm water system nodal depths using auto-regressive graph neural networks,'' \emph{Water Research}, vol. 266, p. 122396, 2024.

\bibitem{fu_making_2024}
G.~Fu, D.~Savic, and D.~Butler, ``Making waves: Towards data-centric water engineering,'' \emph{Water Research}, vol. 256, p. 121585, 2024.

\bibitem{kerimov_towards_2024}
B.~Kerimov, R.~Taormina, and F.~Tscheikner-Gratl, ``Towards transferable metamodels for water distribution systems with edge-based graph neural networks,'' \emph{Water Research}, vol. 261, p. 121933, 2024.

\end{thebibliography}
%%%---------------------------------------------------------------------

%%%---------------------------------------------------------------------
% appendices
\newpage
%\appendices
%%%---------------------------------------------------------------------

%%%---------------------------------------------------------------------
% acknowledgments
% use section* for acknowledgment
%\ifCLASSOPTIONcompsoc
  % The Computer Society usually uses the plural form
%  \section*{Acknowledgments}
%\else
  % regular IEEE prefers the singular form
%  \section*{Acknowledgment}
%\fi
%%%---------------------------------------------------------------------

\ifCLASSOPTIONcaptionsoff
  \newpage
\fi

\end{document}